\documentclass[10pt,twocolumn,letterpaper]{article}

\usepackage{iccv}              

\usepackage[accsupp]{axessibility}
\usepackage{graphicx}
\usepackage{booktabs}
\usepackage{amsmath}
\usepackage{amssymb}
\usepackage{multirow}
\usepackage{xcolor}
\usepackage{soul}
\usepackage{tabularx}
\usepackage{colortbl}  %
\usepackage{array}
\usepackage{siunitx} %
\usepackage{tcolorbox}  %
\usepackage{listings}
\usepackage{mathrsfs}
\usepackage{wrapfig}
\usepackage{upgreek}
\usepackage{makecell}
\usepackage{vcell}
\usepackage{mwe} %
\usepackage{calc} %
\usepackage{pifont} %
\usepackage{marvosym}   %
\usepackage[pagebackref,breaklinks,colorlinks,allcolors=iccvblue]{hyperref}
\usepackage[capitalize]{cleveref}
\usepackage{acronym}
\usepackage[ruled,vlined]{algorithm2e}

\crefname{algorithm}{Alg.}{Algs.}
\Crefname{algocf}{Algorithm}{Algorithms}
\crefname{section}{Sec.}{Secs.}
\Crefname{section}{Section}{Sections}
\crefname{table}{Tab.}{Tabs.}
\Crefname{table}{Table}{Tables}
\crefname{figure}{Fig.}{Fig.}
\Crefname{figure}{Figure}{Figure}
\definecolor{revision}{RGB}{0,0,255}

\definecolor{iccvblue}{rgb}{0.21,0.49,0.74}
\newcommand{\yes}{\color{blue}{\ding{51}}}
\newcommand{\no}{\color{red}{\ding{55}}}
\definecolor{Gray}{gray}{0.85}
\newcommand{\sota}{state-of-the-art\xspace}
\newcommand{\method}{\mbox{{GWM}}\xspace}
\newcommand{\methodfull}{Gaussian World Model\xspace}

\definecolor{best}{rgb}{0.96, 0.57, 0.58}
\definecolor{second}{rgb}{0.98, 0.78, 0.57}
\definecolor{third}{rgb}{1.0, 1.0, 0.56}

\definecolor{shallowgreen}{RGB}{200, 230, 201} 
\definecolor{shalloworange}{RGB}{255, 224, 178} 
\newcommand{\ourcolor}{shallowgreen}

\acrodef{gwm}[GWM]{Gaussian World Model}
\acrodef{dit}[DiT]{Diffusion Transformer}
\acrodef{vae}[VAE]{Variational Autoencoder}
\acrodef{il}[IL]{Imitation Learning}
\acrodef{rl}[RL]{Reinforcement Learning}
\acrodef{mbrl}[MBRL]{Model-based Reinforcement Learning}
\acrodef{pomdp}[POMDP]{Partially Observable Markov Decision Process}
\acrodef{mdp}[MDP]{Markov Decision Process}
\acrodef{3d}[3D]{three-dimensional}
\acrodef{nerf}[NeRF]{Neural Radiance Field}
\acrodef{sde}[SDE]{stochastic differential equation}
\acrodef{3dgs}[3D-GS]{3D Gaussian Splatting}
\acrodef{fps}[FPS]{Farthest Point Sampling}

\newlength\savewidth

\makeatletter
\renewcommand{\paragraph}{%
  \@startsection{paragraph}{4}{\z@}%
  {1ex plus 0.5ex minus 0.2ex} 
  {-1em}                      
  {\normalfont\normalsize\bfseries} 
}
\makeatother

\usepackage{amsmath,amsfonts,bm}

\def\eqref#1{equation~\ref{#1}}

\def\1{\bm{1}}

\def\rva{{\mathbf{a}}}

\def\rvr{{\mathbf{r}}}
\def\rvs{{\mathbf{s}}}

\def\rvw{{\mathbf{w}}}
\def\rvx{{\mathbf{x}}}
\def\rvy{{\mathbf{y}}}

\def\vd{{\bm{d}}}

\def\vx{{\bm{x}}}

\def\mC{{\bm{C}}}

\def\mG{{\bm{G}}}

\def\mI{{\bm{I}}}

\def\mQ{{\bm{Q}}}

\def\mX{{\bm{X}}}

\DeclareMathAlphabet{\mathsfit}{\encodingdefault}{\sfdefault}{m}{sl}
\SetMathAlphabet{\mathsfit}{bold}{\encodingdefault}{\sfdefault}{bx}{n}

\def\gD{{\mathcal{D}}}

\def\gF{{\mathcal{F}}}

\def\gN{{\mathcal{N}}}

\DeclareMathOperator*{\argmax}{arg\,max}

\def\paperID{3042} 
\def\confName{ICCV}
\def\confYear{2025}

\title{GWM: Towards Scalable Gaussian World Models for Robotic Manipulation}

\author{Guanxing Lu$^{1,2,\star}$, Baoxiong Jia$^{2,\star,\dagger}$, Puhao Li$^{2,\star}$, Yixin Chen$^{2}$ \\ Ziwei Wang$^{3}$, Yansong Tang$^{1,\dagger}$, Siyuan Huang$^{2,\dagger}$\\
$^\star$Equal contribution \quad $^\dagger$Corresponding author\\
$^1$ Tsinghua University, $^2$ State Key Laboratory of General Artificial Intelligence, BIGAI\\
$^3$ School of Electrical and Electronic Engineering, Nanyang Technological University\\
\href{https://gaussian-world-model.github.io}{\texttt{\textbf{gaussian-world-model.github.io}}}
\vspace{0.3cm}
}

\begin{document}

\twocolumn[{
\renewcommand\twocolumn[1][]{#1}%
\maketitle
\begin{center}
    \centering
    \includegraphics[width=1\linewidth]{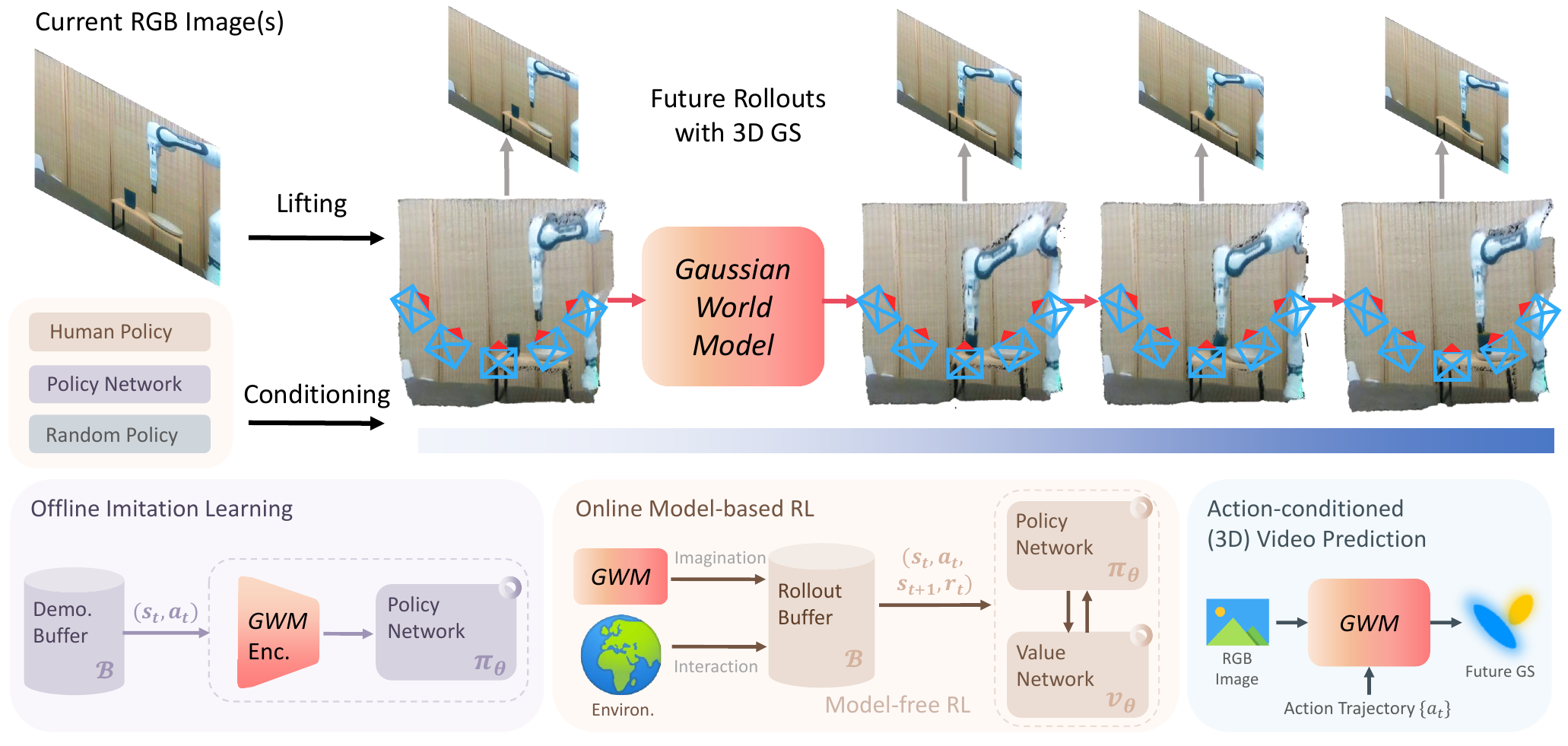}
    \captionsetup{type=figure}
        \captionof{figure}{\textbf{Gaussian World Model (GWM)} is a novel branch of world model that predicts dynamic future states and enables robotic manipulation based on the 3D Gaussian Splatting representation. It facilitates action-conditioned 3D video prediction, enhances visual representation learning for imitation learning, and serves as a robust neural simulator for model-based reinforcement learning.}
\label{fig:overview}
\end{center}%
}]

\begin{abstract}

Training robot policies within a learned world model is trending due to the inefficiency of real-world interactions. The established image-based world models and policies have shown prior success, but lack robust geometric information that requires consistent spatial and physical understanding of the three-dimensional world, even pre-trained on internet-scale video sources. To this end, we propose a novel branch of world model named Gaussian World Model (GWM) for robotic manipulation, which reconstructs the future state by inferring the propagation of Gaussian primitives under the effect of robot actions. At its core is a latent Diffusion Transformer (DiT) combined with a 3D variational autoencoder, enabling fine-grained scene-level future state reconstruction with Gaussian Splatting. GWM can not only enhance the visual representation for imitation learning agent by self-supervised future prediction training, but can serve as a neural simulator that supports model-based reinforcement learning. Both simulated and real-world experiments depict that GWM can precisely predict future scenes conditioned on diverse robot actions, and can be further utilized to train policies that outperform the state-of-the-art by impressive margins, showcasing the initial data scaling potential of 3D world model.

\end{abstract}



\section{Introduction}




Humans construct predictive \emph{world models} from limited sensory input, allowing them to anticipate future outcomes and adapt to new situations~\cite{forrester1971counterintuitive,ha2018recurrent}. Inspired by this capability, world model learning has driven major advances in intelligent agents, enabling strong performance in autonomous driving~\cite{wang2023drivedreamer,gao2022enhance,hu2022model,hu2023gaia,zheng2023occworld} and gaming~\cite{ha2018recurrent,hafner2022deep,hafner2019dream,hafner2020mastering,hafner2023mastering,ye2021mastering,schrittwieser2020mastering,alonso2024diffusion}. As intelligent agents increasingly engage with the physical world, advancing world model learning for robotic manipulation becomes an essential research direction, as it could ideally empower robots to reason about interactions, predict physical dynamics, and adapt to diverse unseen environments.

This naturally raises the following question:  \textit{How to effectively represent, construct, and leverage the world model to enhance robotic manipulation?} Such demand poses significant challenges to existing representations and models.

\begin{itemize}[leftmargin=*]
\item \textbf{Necessity of 3D Representation} 
High-capacity architectures~\cite{vaswani2017attention, ho2020denoising} and internet-scale pre-training have established video-based generative models as powerful tools for capturing world dynamics information, which significantly enhances policy learning~\cite{yang2023learning, wu2025ivideogpt}. 
However, their reliance on image inputs makes them susceptible to unseen visual variations (\eg, lighting, camera pose, textures, \etc)~\cite{li2024evaluating}, as they lack 3D geometric and spatial understanding. While RGB-D and multi-view~\cite{goyal2023rvt,goyal2024rvt2} setups attempt to mitigate this gap, implicitly aligning image patch features within a coherent 3D space remains challenging~\cite{peri2024point,zhu2025point}, leaving the robustness concern unresolved. This \textit{underscores the need for representations that integrate fine visual details with 3D spatial information} to enhance world modeling for robotic manipulation.

\item \textbf{Efficiency and Scalability} 
To identify a 3D representation that preserves both 3D geometric structure and fine visual details from 2D images, multi-view 3D reconstruction methods such as \ac{nerf}~\cite{mildenhall2021nerf} and \ac{3dgs}~\cite{kerbl20233d} offer natural solutions. 
Among them, \ac{3dgs} is particularly appealing due to its explicit per-Gaussian modeling of 3D scenes, marrying efficient 3D representations like point clouds with high-fidelity rendering. However, since these methods primarily rely on offline per-scene reconstruction, their computational demands pose significant challenges~\cite{ze2023gnfactor,lu2024manigaussian} on applying them in robotic manipulation, especially for \ac{mbrl}, limiting their scalability.
\end{itemize}

To this end, we propose \textbf{\methodfull (\method)}, a novel 3D world model that integrates \ac{3dgs} with high-capacity generative models for robotic manipulation. Specifically, our approach combines recent advancements in feed-forward \ac{3dgs} reconstruction with \acp{dit}, enabling fine-grained future scene reconstruction through Gaussian rendering conditioned on current observations and robot actions. To achieve real-time training and inference, we design a 3D Gaussian \ac{vae} to extract latent representations from 3D Gaussians, allowing the diffusion-based world model to operate efficiently in a compact latent space. With this novel design, we demonstrate that \method enhances visual representation learning, improving its role as a visual encoder for imitation learning while also serving as a robust neural simulator for model-based \ac{rl}.

To comprehensively evaluate \method, we conduct extensive experiments in action-conditioned video prediction, imitation learning, and model-based RL settings, covering 31 diverse robotic tasks across 3 domains. For real-world evaluation, we introduce a Franka PnP task suite with 20 variations, encompassing both in-domain and out-of-domain settings. For the ablation study, we evaluate both perceptual metrics and success rates to verify the effectiveness of each building blocks. \method consistently outperforms previous baselines, including state-of-the-art image-based world models, offering notable advantages and highlighting its data-scaling potential.

In summary, our main contributions are threefold. 
\begin{itemize}
    \item We introduce \method, a novel 3D world model that is instantiated with a Gaussian diffusion transformer and a Gaussian \ac{vae} for efficient dynamic modeling. \method learns to predict accurate future states and dynamics in a scalable end-to-end manner without human intervention.
    \item \method can be easily integrated into offline imitation learning and online reinforcement learning with superior efficiency, depicting impressive scaling potential in learning-based robotic manipulation.
    \item We demonstrate the efficacy of \method through extensive experiments in two challenging simulation environments, which improves the previous state-of-the-art baselines by a large margin of $16.25$\%. Furthermore, we validate its practicality in real-world scenarios, where \method improves a typical diffusion policy by $30$\% with $20$ trials.
\end{itemize}


\section{Related Work}
\label{sec:related_work}

\paragraph{World Models} 
World models capture scene dynamics and enable efficient learning by predicting future states based on current observations and actions. They have been widely explored in autonomous driving~\cite{wang2023drivedreamer,gao2022enhance,hu2022model,hu2023gaia,zheng2023occworld,zuo2025gaussianworld}, game agents \cite{ha2018recurrent,hafner2022deep,hafner2019dream,hafner2020mastering,hafner2023mastering,ye2021mastering,schrittwieser2020mastering,alonso2024diffusion}, and robotic manipulation \cite{hansen2023td,wu2023daydreamer,seo2023masked}. 
Early works~\cite{hafner2023mastering,iris2023,robine2023transformer,zhang2023storm,ha2018recurrent,hafner2022deep,hafner2019dream,hafner2020mastering,hansen2023td,ye2021mastering,schrittwieser2020mastering,seo2023masked} learn a latent space for future prediction, achieving strong results in both simulated and real-world settings \citep{wu2023daydreamer}. 
However, while simplifying modeling, latent representations struggle to capture the world's fine details. Recent advances in diffusion models \citep{sohl2015difforigin,ho2020DDPM,song_sde} and transformers\citep{vaswani2017attention, radford2018improving} have shifted world modeling toward direct pixel-space modeling~\citep{yang2023learning, alonso2024diffusion, lu2025dreamart, lu2025taco}, enabling fine-grained detail capture and large scale learning from internet videos. Yet, image-based models often lack physical commonsense~\cite{bansal2024videophy}, thus limiting their applicability in robotic manipulation.


\begin{figure*}[t]
    \centering
    \includegraphics[width=1\textwidth]{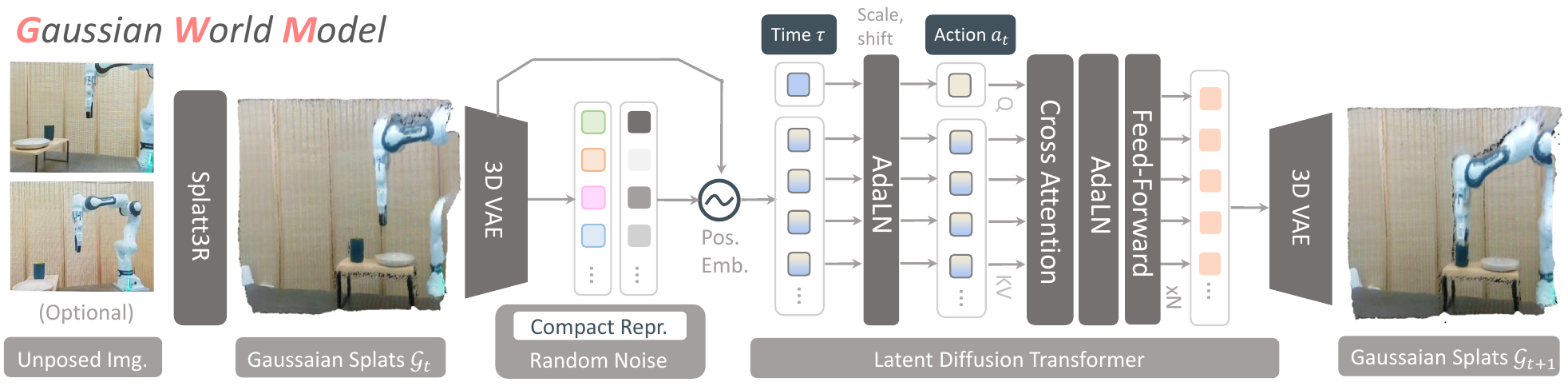}
    \caption{\small \textbf{The overall pipeline of \method}, which primarily consists of a 3D variational encoder and a latent diffusion transformer. The 3D variational encoder embeds the Gaussian Splats estimated by a foundational reconstruction model to a compact latent space, and the diffusion transformer operates on the latent patches to interactively imagine the future Gaussian Splats conditioned on the robot action and denoising time step.
    }
    \label{fig:pipeline}
\end{figure*}

\paragraph{Gaussian Splatting}
\ac{3dgs} \citep{kerbl20233d} represents scenes using 3D Gaussians, which are efficiently projected onto 2D planes via differentiable splitting. Compared to implicit representations like \ac{nerf}~\cite{mildenhall2021nerf}, it offers greater efficiency, benefiting applications such as invasive surgery \citep{liu2024endogaussian}, SLAM \citep{keetha2024splatam}, and autonomous driving \citep{zhou2024drivinggaussian}.
This advantage extends to 4D dynamic modeling~\citep{luiten2024dynamic,huang2024sc,xie2024physgaussian} as the 3D Gaussians, similar to point clouds, are spatially meaningful. However, the offline per-scene reconstruction required by these methods imposes computational challenges for real-time applications like robotic manipulation. Recent works~\citep{szymanowicz2023splatter,xu2024grm, zheng2023gps,zou2023triplane,charatan2023pixelsplat,fu2023colmap,xu2024agg,zhang2024gaussiancube} address this issue by learning generative mappings from pixels to Gaussians using large-scale datasets, but still rely on known camera poses, restricting scalability. A parallel effort~\citep{smart2024splatt3r,wang2024dust3r,leroy2024grounding,chen2023ssr} explores feed-forward novel-view synthesis from unposed images, leveraging the predicted point map as a proxy for explicit multi-view alignment. Building on these advances, this work develops a scalable Gaussian world model from unposed images, ensuring spatial awareness and scalability for policy training.

\paragraph{Visual Manipulation}

Building vision-driven robots with human-like capabilities is a long-standing challenge. Visual imitation learning methods~\citep{liu2024rdt,black2024pi_0,kim2024openvla,team2024octo,li2025controlvla} mimics expert demonstrations using various visual representations, such as point clouds \citep{chen2023polarnet, gervet2023act3d}, voxels \citep{shridhar2023peract,liu2024voxact}, \acp{nerf}~\citep{jiang2021giga,lin2023mira,li20223d,driess2022nerfrl,shim2023snerl,ze2023gnfactor}, and \ac{3dgs} \cite{lu2024manigaussian}. While effective for learned tasks, these models struggle in unseen real-world scenarios~\cite{luo2024serl,luo2024precise}.
\ac{rl} fills in this gap by refining policies through trial and error but requires costly real-world rollouts. Therefore, many methods adopt sim-to-real transfer, \ie, deploying \ac{rl} policies learned in digital twins of the world for task execution. Nonetheless, scalability remains a challenge due to their reliance on predefined assets~\citep{mu2024robotwin,Genesis,wang2023gensim} or labor-intensive conversion of real-world objects into simulation~\citep{lou2024robo,qureshi2024splatsim,dai2024acdc,lim2022real2sim2real,li2024ag2manip,liu2025building}. To address limitations, \method focuses on providing both a stronger visual representation for imitation learning and an efficient neural simulator for visual \ac{rl} to enable more effective and scalable robotic manipulation.








\section{\methodfull}
\label{sec:approach}

The overall pipeline of our \method method is shown in \Cref{fig:pipeline}, in which we construct a Gaussian world model to infer the future scene reconstruction represented by 3D Gaussian primitives. Specifically, we encode the real-world vision inputs into latent 3D Gaussian representations (\cref{subsec:encoding}) and leverage a diffusion-based conditional generative model to learn the dynamics over representations given robot states and actions (\cref{subsec:dynamics modeling}). We demonstrate that \method can be flexibly integrated into both offline imitation learning and online model-based reinforcement learning for diverse robotic manipulation tasks (\cref{subsec:applications}).

\subsection{World State Encoding}\label{subsec:encoding}


\paragraph{Feed-forward 3D Gaussian Splatting}
Given single or two-view image inputs $\mathcal{I} = \{\mI\}_{i=\{1,2\}}$ of a world state, our goal is to first encode the scene into 3D Gaussian representations for dynamics learning and prediction. \ac{3dgs} represents a 3D scene with multiple unstructured 3D Gaussian kernels $\mG=\{\vx_p, \bm{\sigma}_p, \bm \Sigma_p, \mathcal{C}_p\}_{p\in \mathcal{P}}$, where $\vx_p$, $\bm{\sigma}_p$, $\bm \Sigma_p$, and $\mathcal{C}_p$ represent the centers, opacities, covariance matrices, and spherical harmonic coefficients of the Gaussians, respectively. To obtain the color of each pixel from a given viewpoint, \ac{3dgs} projects the 3D Gaussians onto the image plane and computes the pixel color as:
\begin{equation}
    \bm C(\mG) = \sum_{p\in \mathcal{P}} \alpha_p \text{SH}(\vd_p; \mathcal{C}_p)  \prod_{j=1}^{p-1} (1-\alpha_j),
\end{equation}
where $\alpha_p$ represents the $z$-depth ordered effective opacities, \ie, products of the 2D Gaussian weights derived from $\bm \Sigma_p$ and their overall opacities $\bm \sigma_p$; $\vd_p$ stands for the view direction from the camera to $\vx_p$; $\text{SH}(\cdot)$ is the spherical harmonics function. Since vanilla \ac{3dgs} relies on time-consuming per-scene offline optimization, we employ generalizable \ac{3dgs} to learn feed-forward mappings from images to 3D Gaussians to accelerate the process. Specifically, we obtain the 3D Gaussian world state $\mG$ using Splatt3R~\cite{smart2024splatt3r}, which first employs the stereo reconstruction model Mast3R~\cite{leroy2024grounding} to generate 3D point maps from input images and then predicts the parameters of each 3D Gaussian given these point maps using an additional prediction head.

\paragraph{3D Gaussian VAE} 
Since the number of learned 3D Gaussians for each world state can vary significantly across different scenes and tasks, we adopt a 3D Gaussian VAE $(E_\theta,D_\theta)$ to encode the reconstructed 3D Gaussians $\mG$ into a fixed length of $N$ latent embeddings $\rvx\in\mathbb{R}^{N\times D}$. Specifically, we first downsample the reconstructed 3D Gaussians $\mG$ to a fixed number of $N$ Gaussians $\mG_{N}$ using \ac{fps}: $\mG_{N} = \text{FPS}(\mG)$. Next, we use these sampled Gaussians $\mG_{N}$ as queries to attend and aggregate information from all Gaussians $\mG$ to latent embedding $\rvx$ using a $L$ layer cross-attention-based encoder $E_\theta$ like \citep{zhang20233dshape2vecset}:
\begin{equation}
\begin{aligned}
    & \mX = E_\theta(\mG_N, \mG) = E_{\theta}^{(L)} \circ \dots \circ E_{\theta}^{(1)} (\mG_N, \mG),\\
    E_{\theta}^{(l)}(\mQ&,\mG) = \text{LayerNorm}(\text{CrossAttn}(\mQ, \text{PosEmbed}(\mG))).
\end{aligned}
\end{equation}
With latent encoding $\rvx$, we employ a mirrored transformer-based decoder $D_\theta$ to propagate and aggregate information within the latent code set and leverage to obtain the reconstructed Gaussians $\hat{\mG}$:
\begin{equation}
\hat{\mG} = D_\theta (\rvx) = \text{LayerNorm}(\text{SelfAttn}(\rvx, \rvx))
\end{equation}


For learning the 3D Gaussian VAE $(E_\theta, D_\theta)$, we use the Chamfer loss between the centers of our reconstructed Gaussians $\hat{\mG}$ and the original ones $\mG$ for supervision. We also add a rendering loss of our reconstructed Gaussians $\hat{\mG}$ to achieve high-fidelity rendering for image-based policy:
\begin{equation}
\mathcal{L}_{\text{VAE}} = \text{Chamfer}(\hat{\mG}, \mG) + \|\mC(\hat{\mG})-\mC(\mG)\|_1
\end{equation}

\subsection{Diffusion-based Dynamics Modeling}\label{subsec:dynamics modeling}
With the encoded world state embeddings $\rvx_t$ at time $t$ and its future state $\rvx_{t+1}$, we aim to learn the world dynamics $p(\rvx_{t+1} | \rvx_{\leq t}, a_{\leq t})$, where $\rvx_{\leq t}$ and $a_{\leq t}$ denote history states and actions, respectively. Specifically, we leverage a diffusion-based dynamics model where we convert dynamics learning into a conditional generation problem, generating future state $\rvx_{t+1}$ from noise with history states and actions $\rvy_{t} = (\rvx_{\leq t}, a_{\leq t})$ as conditions. 

\paragraph{Diffusion Formulation} To generate the future state, we begin with the formulation of the diffusion process. Specifically, we first add noise to the ground truth future state $\rvx_{t+1}^0 = \rvx_{t+1}$ to obtain noised future state samples $\rvx_{t+1}^\tau$ via a Gaussian perturbation kernel:
\begin{equation}
    \label{eq:add_noise}
    p^{0\to\tau}(\rvx_{t+1}^\tau | \rvx_{t+1}^0) = \gN(\rvx_{t+1}^\tau; \rvx_{t+1}^0, \sigma^2(\tau)\mI),
\end{equation}
where $\tau$ is the noise step index and $\sigma(\tau)$ is the noise schedule. This diffusion process can be described as the solution to a \ac{sde}~\cite{song_sde}:
\begin{equation}
\label{eq:forward_process}
    d\rvx = \mathbf{f} (\rvx, \tau) d\tau +  g(\tau) d\rvw, 
\end{equation}
where $\rvw$ represents the standard Wiener process, $\mathbf{f}$ is the drift coefficient, and $g$ is the diffusion coefficient. Under this formulation, the effect of the Gaussian perturbation kernel is equivalent to setting $\mathbf{f}(\rvx, \tau) = \mathbf{0}$ and $g(\tau) = \sqrt{2 \dot \sigma(\tau) \sigma(\tau)}$. To generate samples from noises, we can reverse~\cref{eq:forward_process} using the reverse-time \ac{sde}~\cite{anderson1982reverse} for sampling:
\begin{equation}
\label{eq:reverse_process}
    d\rvx = [\mathbf{f} (\rvx, \tau) - g(\tau)^2 \nabla_\rvx \log p^\tau(\rvx)] d\tau +  g(\tau) d\Bar{\rvw},
\end{equation}
where $\Bar{\rvw}$ denotes the reverse-time Wiener process and $\nabla_\rvx \log p^\tau(\rvx)$ is the score function, \ie, the gradient of log-marginal probability with respect to $\rvx$ \citep{hyvarinen2005estimation}. As the score function could be estimated by a network, we learn the conditional denoising model $\gD_\theta$  by minimizing the difference between the sampled future state $\hat{\rvx}_{t+1}^0 = \gD_\theta(\rvx_{t+1}^\tau, \rvy_t)$ and the ground truth future state $\rvx_{t+1}^0$:
\begin{equation}
    \label{eq:diffusion_ori}
    \mathcal{L}(\theta) = \mathbb{E} \left[ \left\| \gD_\theta(\rvx_{t+1}^\tau, y_t^\tau) - \rvx^{0}_{t+1} \right\|_2^2 \right].
\end{equation}

\paragraph{Learning with EDM} As pointed out in \cite{karras2022elucidating}, directly learning the denoiser $\gD_\theta(\rvx_{t+1}^\tau, \rvy_t)$ can be affected by problems like varying noise magnituides. Therefore, we follow \cite{alonso2024diffusion} and adopt the practice in EDM~\cite{karras2022elucidating} to learn a network $\gF_\theta$ with preconditioning instead. Specifically, we parameterize the denoiser $\gD_{\theta}(\rvx_{t+1}^\tau,\rvy_{t+1}^\tau)$ with:
\begin{equation}
    \gD_\theta(\rvx_{t+1}^\tau, \rvy_t^{\tau}) = c_\text{skip}^\tau \; \rvx_{t+1}^\tau + c_\text{out}^\tau \; \gF_\theta \big( c_\text{in}^\tau \; \rvx_{t+1}^\tau, \rvy_t^\tau; c_{\text{noise}}^\tau\big),
\end{equation}
where the preconditioner $c_\text{in}^\tau$ and $c_{\text{out}}^\tau$ scale the input and output magnitudes, $c_{\text{skip}}^\tau$ modulates the skip connection, and $c_{\text{noise}}^\tau$ maps noise levels as an additional conditioning input into $\gF_{\theta}$. 
We provide details for these preconditioners in~\cref{app:method:edm_precond}. 
With this conversion, we can rewrite the objective in~\cref{eq:diffusion_ori} with:
\begin{equation}
    \label{eq:diffusion_new}
    \mathcal{L}(\theta) = \mathbb{E} \left[ \left\| 
   \mathbf{F}_{\theta} \left( c_{\text{in}}^{\tau} \mathbf{x}_{t+1}^{\tau}, y_t^{\tau} \right)
    - \frac{1}{c_{\text{out}}^{\tau}} \left( \mathbf{x}_{t+1}^{0} - c_{\text{skip}}^{\tau} \mathbf{x}_{t+1}^{\tau} \right)
    \right\|_2^2 \right].
\end{equation}
A crucial insight of this conversion is creating a new training target for better learning the network $\mathcal{F}_\theta$ by adaptively mixing signal and noise depending on the noise schedule $\sigma(\tau)$. Intuitively, at high noise levels ($\sigma(\tau) \gg \sigma_\text{data}$), $c_\text{skip}^\tau \to 0$ and the network primarily learns to predict the clean signal. Conversely, at low noise levels ($\sigma(\tau) \to 0$), $c_\text{skip}^\tau \to 1$, the target becomes the noise component, preventing the objective from becoming trivial.

\paragraph{Implementation} Technically, we implement the network $\gF_\theta$ with a \ac{dit}~\cite{peebles2023scalable}. Given a sequence of actual world state latent embeddings $\{\rvx_t^0 = \rvx_t\}_{t=1}^{T}$, we first create latents with noise $\{\rvx_t^\tau\}_{t=1}^T$ following the Gaussian perturbation described in~\cref{eq:add_noise}. Next, we concat the noise latent embeddings with rotary position embedding (RoPE~\cite{su2024roformer}) and pass it to the \ac{dit} as inputs. In terms of conditions $\rvy_t = (\rvx_{\leq t}^0, a_{\leq t}, c_{\text{noise}}^\tau)$, the time embeddings are modulated by adaptive layer normalization (AdaLN \citep{perez2018film}), and the current robot actions are used as the keys and values for the cross-attention layers within the \ac{dit} for conditional generation. 
For stability and efficiency across all attention mechanisms, we employ Root Mean Square Normalization (RMSNorm \citep{zhang2019root}) with learnable scales to stabilize training that processes spatial representations while incorporating temporal action sequences as conditions.



\subsection{\method for Policy Learning}\label{subsec:applications}

\begin{algorithm}[t]
\caption{Monotonic Model-Based Policy Optimization (MBPO) with Gaussian World Model}
\SetAlgoLined
Initialize policy $\pi(\rva_t|\rvs_t)$, Gaussian world model $p_\theta(\rvs_{t+1}, \rvr_t | s_t, \rva_t)$, empty replay buffer $\mathcal{B}$\;
\For{$N$ epochs}{
    Collect data with $\pi$ in real environment: $\mathcal{B} = \mathcal{B} \cup \{(\rvs_t, \rva_t, \rvs_{t+1}, \rvr_t)\}_t$\;
    Train Gaussian world model $p_\theta$ on dataset $\mathcal{B}$ via maximum likelihood: $\theta \leftarrow \argmax_\theta \mathbb{E}_\mathcal{B}[\log p_\theta(\rvs_{t+1}, \rvr_t|s_t, \rva_t)]$\;
    Optimize policy under predictive model: $\pi \leftarrow \argmax_{\pi} \mathbb{E}_\pi[\sum_{t \ge 0} \gamma^t \rvr_t]$\;
}
\label{alg:mbpo_gwm}
\end{algorithm}

\paragraph{\method for Reinforcement Learning}\label{subsubsec:reinforcement learning} We demonstrate that \method can be seamlessly integrated into existing model-based \ac{rl} methods. Formally, a \ac{mdp} is defined by the tuple $(\mathcal{S}, \mathcal{A}, p, r, \gamma, \rho_0)$. $\mathcal{S}$ and $\mathcal{A}$ are the state and action spaces, $\gamma$ is the discount factor, and $r(\rvs,\rva)$ is the reward function. The goal of model-based \ac{rl}~\cite{janner2021trust} is to learn a policy $\pi$ that maximizes the expected sum of discounted rewards $\pi^* = \arg\max_{\pi} \mathbb{E}_{\pi} \left[ \sum_{t=0}^{\infty} \gamma^t \rvr_t \right]$ while constructing a model of the dynamics $p_\theta(\rvs_{t+1}, \rvr_t |\rvs_t, \rva_t)$ using the policy roll-outs. We provide the pseudo-code for the model-based \ac{rl} policy learning in~\cref{alg:mbpo_gwm}. Under this formulation, we add an additional reward prediction head over \method to parameterize the dynamics model $p_\theta(\rvs_{t+1},\rvr_t | \rvs_t, \rva_t)$. To improve performance in visual manipulation tasks, we build our base \ac{rl} policy following design choices discussed in~\cite{wu2025ivideogpt}.

\paragraph{\method for Imitation Learning}\label{subsubsec:imitation learning}
In imitation learning, we use \method as a more effective encoder to provide better features for policy learning. Specifically, we use the feature vector after the first denoising step in the diffusion process as the input for downstream policy models like BC-transformer~\cite{nasiriany2024robocasa} and diffusion policy~\cite{chi2023diffusion}. 
The first denoising step carries out the representative spatial information to deal with the severe noise level.
In our implementations, we predict actions in sequential chunks to promote consistency in robotic control.

\section{Experiments}
\label{sec:experiments}

In our experiments, we focus on the following questions:
\begin{enumerate}
    \item How is the quality of the action-conditioned video prediction results across different domains?
    \item Does Gaussian world model benefits downstream imitation and reinforcement learning? Does it show greater robustness compared with image-based world model?
    \item How does the Gaussian world model help typical policies (\eg, diffusion policy \citep{chi2023diffusion}) in real-world robotic manipulation tasks?
\end{enumerate}
In the following sections, we describe in detail the model performance regarding these key topics. Specifically, we leverage the following three testing environments and four tasks in our experiments:

\paragraph{Environments}
To provide a comprehensive analysis of \method's capability, we evaluate our method on two synthetic and one real-world environment: (1) \textsc{Meta-World}~\cite{yu2020meta}, a synthetic environment for learning \ac{rl} policies for robotic manipulation; (2) \textsc{RoboCasa}~\cite{nasiriany2024robocasa}, a large-scale multi-scale synthetic imitation learning benchmark featuring diverse robotic manipulation tasks in the kitchen environment; and (3) \textsc{Franka-PnP}, a real-world pick-and-place environment using a Franka Emika FR3 robot arm.

\paragraph{Tasks}
We meticulously design four tasks to evaluate \method across various testing environments systematically: (1) Action-conditioned scene prediction assesses \method's effectiveness in world modeling and future prediction; (2) \method-based imitation learning examines the representation quality and its benefits for imitation-learning-based robotic manipulation; (3) \method-based \ac{rl} explores its potential for model-based reinforcement learning; and (4) real-world task deployment evaluates \method's robustness in real-world robot manipulation.







\subsection{Action-conditioned Scene Prediction}\label{subsec:recon}

\paragraph{Experiment Setup}
The capability of a world model to generate high-fidelity and action-aligned rollouts is critical for effective policy optimization. To evaluate this capability, we train \method on human demonstrations available on all considered real and synthetic environments, and evaluate the future prediction quality by conditioning the model on unseen action trajectories sampled from the validation set. For quantitative evaluation, we employ common metrics for generation quality, including FVD \citep{fvd} to measure temporal consistency, image-based metrics including PSNR \citep{psnr} for pixel-level accuracy, alongside SSIM \citep{ssim} and LPIPS \citep{lpips} for perceptual quality.

\begin{table}[t]
    \caption{\textbf{Quantitative results for future state prediction} on Meta-World and \textsc{Franka PnP}. LPIPS and SSIM scores are scaled by 100. Best results are highlighted in bold.}
    \label{tab:video_pred}
    \centering
    \resizebox{\linewidth}{!}{
        \begin{tabular}{llcccc}
        \toprule
        \textbf{Dataset} & \textbf{Method} & FVD$\downarrow$ & PSNR$\uparrow$ & SSIM$\uparrow$ & LPIPS$\downarrow$ \\ 
        \midrule
        \multirow{2}{*}{\textsc{Meta-World}} & iVideoGPT & 75.0 & 20.4 & 82.3 & 9.5 \\
        & \method & \textbf{73.0} & \textbf{20.6} & \textbf{82.8} & \textbf{9.0} \\
        \midrule
        \multirow{2}{*}{\textsc{Franka-PnP}} & iVideoGPT & 63.2 & 27.8 & 90.6 & 4.9 \\
        & \method & \textbf{61.5} & \textbf{28.0} & \textbf{91.0} & \textbf{4.5} \\
        \bottomrule
        \end{tabular}
    }
\end{table}

\begin{figure}[t]
    \centering
    \includegraphics[width=0.48\textwidth]{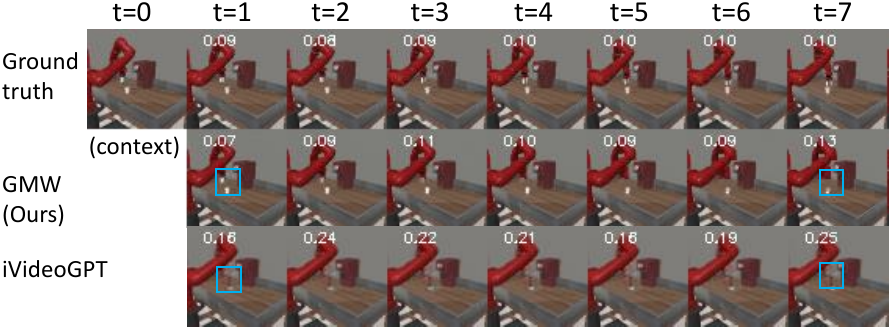}
    \caption{\textbf{Qualitative comparison between models on \textsc{MetaWorld}.} \method successfully predicts better details on the gripper movement (highlighted in blue).
     }
    \label{fig:video-prediction-comparison-results}
\end{figure}

\begin{figure}[t]
    \centering
    \includegraphics[width=0.48\textwidth]{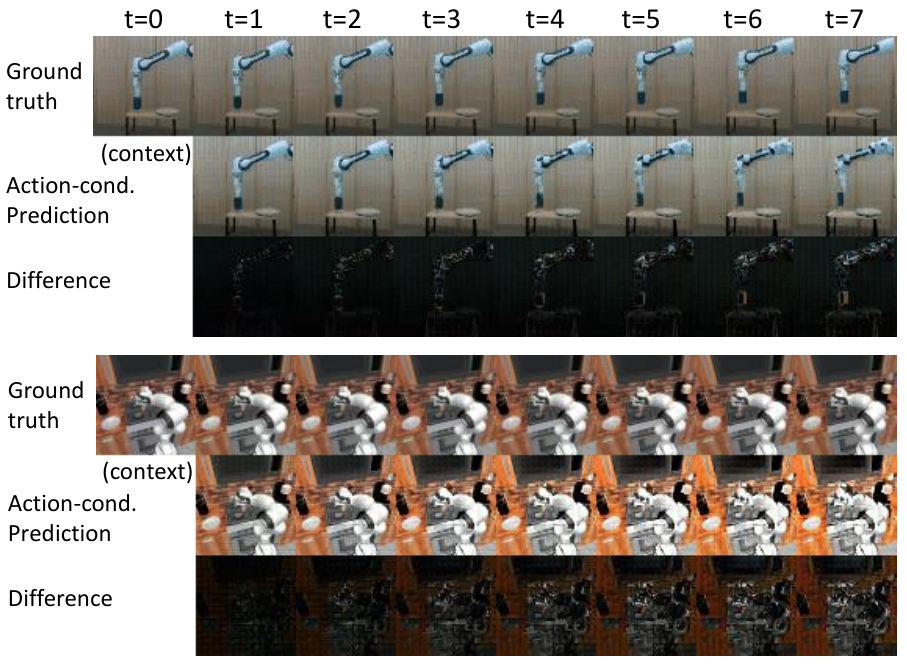}
    \caption{\small \textbf{Qualitative visualization on future state prediction of \method on \textsc{Franka-PnP} and \textsc{RoboCasa}.} All predictions are rolled out by applying the unseen action trajectory from the valid dataset. Zoom in for details. 
     }
    \label{fig:video-prediction-results}
\end{figure}

\paragraph{Results and Analyses} 
We provide quantitative comparison between our method and iVideoGPT in~\cref{tab:video_pred}. As shown in~\cref{tab:video_pred}, our method consistently outperforms the current \sota image-based world modeling method iVideoGPT on both synthetic and real-world environments, demonstrating the effectiveness of our diffusion-based Gaussian world model learning pipeline. Notably, as shown in~\cref{fig:video-prediction-comparison-results}, image-based models like iVideoGPT are prone to failures in capturing dynamics details (\eg, the gripper). Though these details might not cause large differences in visual metrics, they will significantly affect policy learning as we later discuss in~\cref{subsec:rl}. We provide more qualitative visualizations of \method's prediction result on \textsc{RoboCasa} and \textsc{Franka-PnP} in~\cref{fig:video-prediction-results}.


\subsection{\method-based Imitation Learning}\label{subsec:imitate}

\definecolor{flamingoRed}{RGB}{255, 99, 71}
\definecolor{iceBlue}{RGB}{135, 206, 235}

\begin{table*}[t]
\centering
\scriptsize
\caption{\small  \textbf{Multi-Task Imitation Learning Results in Robocasa.} Average success rates (\%) of multi-task agents trained with $50$ human demonstrations or $3000$ generated demonstrations per task. Results are evaluated over $50$ episodes with different floor plans and styles.
}
\setlength{\tabcolsep}{4pt}
\begin{tabular}{l|cccccccccccc} 
\toprule
& \multicolumn{2}{c}{\makecell{\texttt{PnP}\\\texttt{CabToCounter}}}     & \multicolumn{2}{c}{\makecell{\texttt{PnP}\\\texttt{CounterToCab}}} &  \multicolumn{2}{c}{\makecell{\texttt{PnP}\\\texttt{CounterToMicrowave}}}   &
\multicolumn{2}{c}{\makecell{\texttt{PnP}\\\texttt{CounterToSink}}} &
\multicolumn{2}{c}{\makecell{\texttt{PnP}\\\texttt{CounterToStove}}}   & 
\multicolumn{2}{c}{\makecell{\texttt{PnP}\\\texttt{MicrowaveToCounter}}} 
 
\\
\cmidrule(lr){2-3} \cmidrule(lr){4-5} \cmidrule(lr){6-7} \cmidrule(lr){8-9} \cmidrule(lr){10-11} \cmidrule(lr){12-13} 

\\[-13pt]                                                         \\
\vcell{Method}  & \vcell{H-50}  & \vcell{G-3000}  & \vcell{H-50}  & \vcell{G-3000}  & \vcell{H-50}  & \vcell{G-3000}  & \vcell{H-50}  & \vcell{G-3000}  & \vcell{H-50}  & \vcell{G-3000}  & \vcell{H-50}  & \vcell{G-3000} \\[-\rowheight] 
\printcellbottom  & \printcellbottom & \printcellbottom  & \printcellbottom & \printcellbottom  & \printcellbottom & \printcellbottom   & \printcellbottom & \printcellbottom   & \printcellbottom & \printcellbottom   & \printcellbottom  & \printcellbottom  \\[0pt] 
\midrule
BC-transformer & 2  & 18 & 6  & 28 & 2 & 18 & 2 & 44 & 2 & 6 & 2 & 8 \\
\rowcolor{\ourcolor}
\textbf{\method}  & 18  & 32 &  4 & 22 & 14 & 44 & 20 & 38 & 2 & 18 & 20 & 26 \\
$\Delta$  & \textcolor{flamingoRed}{+16}  & \textcolor{flamingoRed}{+14} &  \textcolor{iceBlue}{-2} & \textcolor{iceBlue}{-6} & \textcolor{flamingoRed}{+12} & \textcolor{flamingoRed}{+26} & \textcolor{flamingoRed}{+18} & \textcolor{iceBlue}{-6} & \textcolor{flamingoRed}{0} & \textcolor{flamingoRed}{+12} & \textcolor{flamingoRed}{+18} & \textcolor{flamingoRed}{+18} \\
\midrule
& \multicolumn{2}{c}{\makecell{\texttt{PnP}\\\texttt{SinkToCounter}}} &
\multicolumn{2}{c}{\makecell{\texttt{PnP}\\\texttt{StoveToCounter}}} & \multicolumn{2}{c}{\makecell{\texttt{Open}\\\texttt{SingleDoor}}}      & \multicolumn{2}{c}{\makecell{\texttt{Open}\\\texttt{DoubleDoor}}} & 
\multicolumn{2}{c}{\makecell{\texttt{Close}\\\texttt{DoubleDoor}}} &
\multicolumn{2}{c}{\makecell{\texttt{Close}\\\texttt{SingleDoor}}}
\\
\cmidrule(lr){2-3} \cmidrule(lr){4-5} \cmidrule(lr){6-7} \cmidrule(lr){8-9} \cmidrule(lr){10-11} \cmidrule(lr){12-13} 
\\[-6pt]          
\vcell{} & \vcell{H-50} & \vcell{G-3000} & \vcell{H-50} & \vcell{G-3000} & \vcell{H-50} & \vcell{G-3000} & \vcell{H-50} & \vcell{G-3000} & \vcell{H-50} & \vcell{G-3000} & \vcell{H-50} & \vcell{G-3000}  \\[-\rowheight]
\printcellbottom & \printcellbottom & \printcellbottom & \printcellbottom & \printcellbottom & \printcellbottom & \printcellbottom & \printcellbottom & \printcellbottom & \printcellbottom & \printcellbottom & \printcellbottom & \printcellbottom \\[1pt]
\midrule
BC-transformer & 8  & 42 & 6  & 28 & 46 & 50 & 28 & 48 & 28 & 46 & 56 & 94 \\
\rowcolor{\ourcolor}
\textbf{\method}  & 22  & 38 &  18 & 44 & 58 & 62 & 28 & 42 & 50 & 58 & 54 & 90 \\
$\Delta$  & \textcolor{flamingoRed}{+14}  & \textcolor{iceBlue}{-4} &  \textcolor{flamingoRed}{+12} & \textcolor{flamingoRed}{+16} & \textcolor{flamingoRed}{+12} & \textcolor{flamingoRed}{+12} & \textcolor{flamingoRed}{0} & \textcolor{iceBlue}{-6} & \textcolor{flamingoRed}{+22} & \textcolor{flamingoRed}{+12} & \textcolor{iceBlue}{-2} & \textcolor{iceBlue}{-4} \\
\midrule
& \multicolumn{2}{c}{\makecell{\texttt{Open}\\\texttt{Drawer}}} &
\multicolumn{2}{c}{\makecell{\texttt{Close}\\\texttt{Drawer}}} & \multicolumn{2}{c}{\makecell{\texttt{TurnOn}\\\texttt{Stove}}}      & \multicolumn{2}{c}{\makecell{\texttt{TurnOff}\\\texttt{Stove}}} & 
\multicolumn{2}{c}{\makecell{\texttt{TurnOn}\\\texttt{SinkFaucet}}} &
\multicolumn{2}{c}{\makecell{\texttt{TurnOff}\\\texttt{SinkFaucet}}}
\\
\cmidrule(lr){2-3} \cmidrule(lr){4-5} \cmidrule(lr){6-7} \cmidrule(lr){8-9} \cmidrule(lr){10-11} \cmidrule(lr){12-13} 
\\[-6pt]          
\vcell{} & \vcell{H-50} & \vcell{G-3000} & \vcell{H-50} & \vcell{G-3000} & \vcell{H-50} & \vcell{G-3000} & \vcell{H-50} & \vcell{G-3000} & \vcell{H-50} & \vcell{G-3000} & \vcell{H-50} & \vcell{G-3000}  \\[-\rowheight]
\printcellbottom & \printcellbottom & \printcellbottom & \printcellbottom & \printcellbottom & \printcellbottom & \printcellbottom & \printcellbottom & \printcellbottom & \printcellbottom & \printcellbottom & \printcellbottom & \printcellbottom \\[1pt]
\midrule
BC-transformer & 42  & 74 & 80  & 96 & 32 & 46 & 4 & 24 & 38 & 34 & 50 & 72 \\
\rowcolor{\ourcolor}
\textbf{\method}  & 56  & 90 & 80 & 90 & 46 & 80 & 22 & 40 & 52 & 48 & 44 & 66 \\
$\Delta$  & \textcolor{flamingoRed}{+14}  & \textcolor{flamingoRed}{+16} &  \textcolor{flamingoRed}{0} & \textcolor{iceBlue}{-6} & \textcolor{flamingoRed}{+14} & \textcolor{flamingoRed}{+24} & \textcolor{flamingoRed}{+18} & \textcolor{flamingoRed}{+16} & \textcolor{flamingoRed}{+14} & \textcolor{flamingoRed}{+14} & \textcolor{iceBlue}{-6} & \textcolor{iceBlue}{-6} \\
\midrule
& \multicolumn{2}{c}{\makecell{\texttt{Turn}\\\texttt{SinkSpout}}} &
\multicolumn{2}{c}{\makecell{\texttt{CoffeePress}\\\texttt{Button}}} & \multicolumn{2}{c}{\makecell{\texttt{TurnOn}\\\texttt{Microwave}}}      & \multicolumn{2}{c}{\makecell{\texttt{TurnOff}\\\texttt{Microwave}}} & 
\multicolumn{2}{c}{\makecell{\texttt{CoffeeServe}\\\texttt{Mug}}} &
\multicolumn{2}{c}{\makecell{\texttt{CoffeeSetup}\\\texttt{Mug}}}
\\
\cmidrule(lr){2-3} \cmidrule(lr){4-5} \cmidrule(lr){6-7} \cmidrule(lr){8-9} \cmidrule(lr){10-11} \cmidrule(lr){12-13} 
\\[-6pt]          
\vcell{} & \vcell{H-50} & \vcell{G-3000} & \vcell{H-50} & \vcell{G-3000} & \vcell{H-50} & \vcell{G-3000} & \vcell{H-50} & \vcell{G-3000} & \vcell{H-50} & \vcell{G-3000} & \vcell{H-50} & \vcell{G-3000}  \\[-\rowheight]
\printcellbottom & \printcellbottom & \printcellbottom & \printcellbottom & \printcellbottom & \printcellbottom & \printcellbottom & \printcellbottom & \printcellbottom & \printcellbottom & \printcellbottom & \printcellbottom & \printcellbottom \\[1pt]
\midrule
BC-transformer & 54  & 96 & 48  & 74 & 62 & 90 & 70 & 60 & 22 & 34 & 0 & 12 \\
\rowcolor{\ourcolor}
\textbf{\method}  & 72  & 90 & 76 & 90 & 64 & 84 & 70 & 54 & 36 & 50 & 16 & 28 \\
$\Delta$  & \textcolor{flamingoRed}{+18}  & \textcolor{iceBlue}{-6} &  \textcolor{flamingoRed}{+28} & \textcolor{flamingoRed}{+16} & \textcolor{flamingoRed}{+2} & \textcolor{iceBlue}{-6} & \textcolor{flamingoRed}{0} & \textcolor{iceBlue}{-6} & \textcolor{flamingoRed}{+14} & \textcolor{flamingoRed}{+16} & \textcolor{flamingoRed}{+16} & \textcolor{flamingoRed}{+16} \\
\bottomrule
\end{tabular}
\vspace{-0.1cm}
\label{table:robocasa-results}
\end{table*}


\paragraph{Experiment Setup}
As discussed in \cref{subsubsec:imitation learning}, \method can be used to extract informative representation from image observation, which is expected to benefit imitation learning.
We verify this property by testing \method's effectiveness for imitation learning on \textsc{RoboCasa}. The task suite in \textsc{RoboCasa} comprises 24 atomic tasks with related language instructions for kitchen environments, including actions such as pick-and-place, open, and close. Each task is provided with a limited set of $50$ human demonstrations and a set of $3000$ generated demonstrations from MimicGen~\citep{mandlekar2023mimicgen}. We train our \method on these demonstrations and pass it as the state encoding for the state-of-the-art BC-transformer~\cite{nasiriany2024robocasa} for quantitative comparison on the success rate metrics.

\paragraph{Results and Analyses} 
Our experimental results on the \textsc{Robocasa} benchmark are presented in Table \ref{table:robocasa-results}, which demonstrate the effectiveness of our method in multi-task imitation learning scenarios. Across $24$ kitchen manipulation tasks, our approach consistently outperforms the BC-Transformer baseline. With limited human demonstrations (H-50), our method shows an average improvement of $10.5\%$ in success rates. When trained on generated demonstrations (G-3000), our method maintains scalable performance with an average gain of $7.6\%$. Notably, our approach exhibits particular strengths in complex manipulation tasks such as pick-and-place operations and interactive tasks like turning on/off appliances, where the performance gains are most significant. 
These results confirm that our method's ability to extract informative representations from visual observations effectively enhances imitation learning capabilities in practical robotic manipulation scenarios.

\subsection{\method-based Reinforcement Learning}\label{subsec:rl}

\begin{figure*}[t]
    \centering
    \includegraphics[width=1\textwidth]{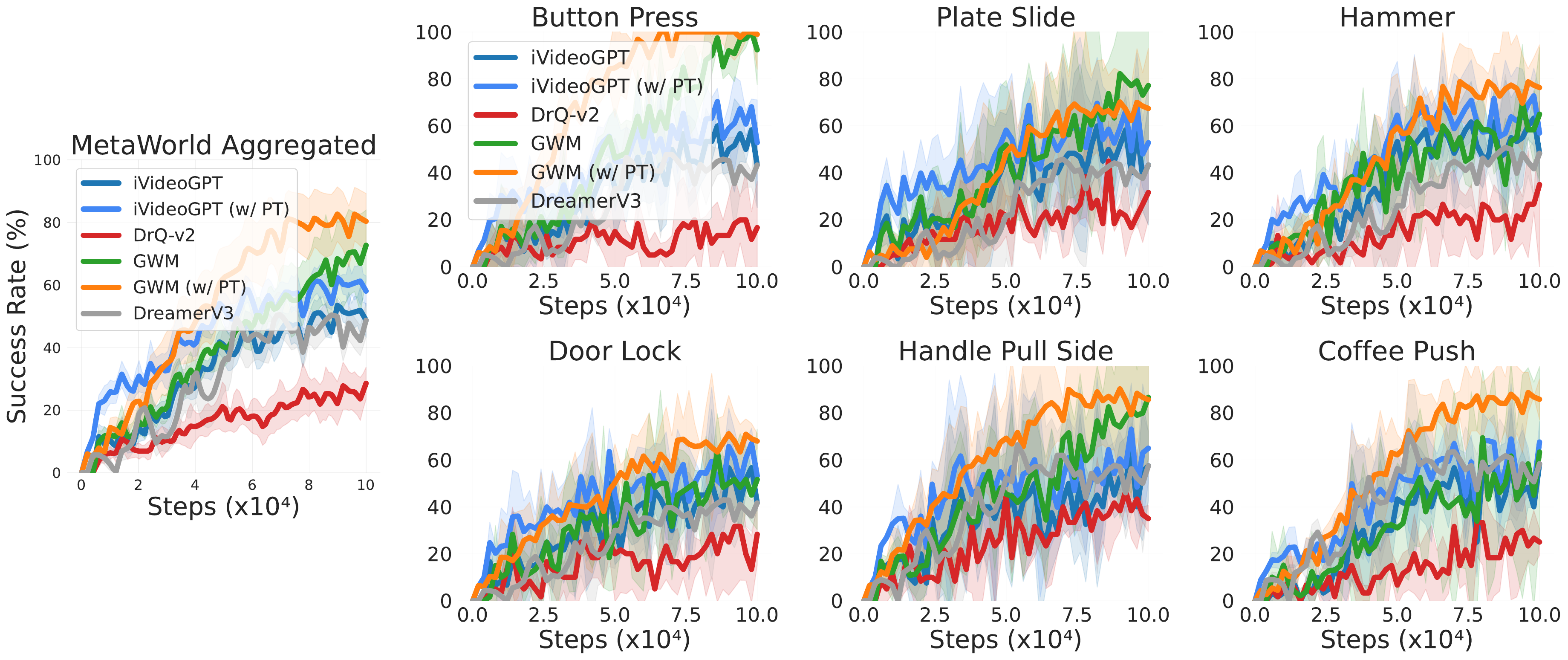}
    \caption{\small \textbf{Model-based RL Results of \method and ivideogpt \citep{wu2025ivideogpt} on \textsc{Metaworld}.} The shadow area represents $95$\% confidence interval (CI) across three random seeds. Each data point is evaluated over $20$ episodes.
    }
    \label{fig:model-based-curves}
\end{figure*}


\paragraph{Experiment Setup}
We evaluate \method's capabilities for \ac{rl} policies on six Meta-World \cite{yu2020meta} robotic manipulation tasks with increasing complexity. We implement a model-based RL approach inspired by MBPO \cite{janner2021trust}, using \method to generate synthetic rollouts that augment the replay buffer of a DrQ-v2 \cite{yarats2021mastering} actor-critic algorithm. We include the \sota image-based world model iVideoGPT \citep{wu2025ivideogpt} as a strong baseline. For fair comparisons, we do not utilize pre-trained initialization of both methods.
For fair comparisons, all compared methods use the same context length, horizon, and are trained to a maximum of $1\times10^5$ steps.

\paragraph{Results and Analyses}
Figure \ref{fig:model-based-curves} demonstrates that GWM consistently outperforms iVideoGPT across all six Meta-World tasks. 
On average, GWM converges approximately $2\times$ faster than iVideoGPT and reaches higher asymptotic performance on complex manipulation tasks.
The superior performance stems from GWM's 3D Gaussian representation, which allows more accurate prediction of contact dynamics and object movement under manipulation, compared to purely image-based approaches. 
The results confirm that explicit 3D representation offers substantial advantages for robotic control tasks requiring precise spatial reasoning.


\begin{figure}[t]
    \centering
    \includegraphics[width=0.45\textwidth]{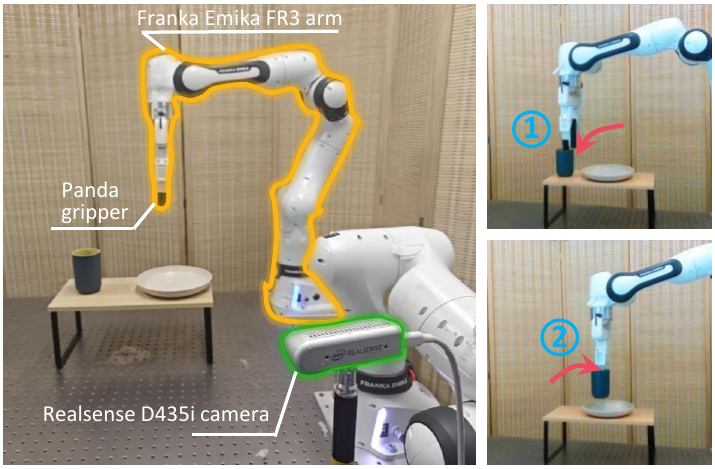}
    \caption{\small \textbf{Real-World Experiment Setup.} Left: using a Franka Emika Panda robotic arm equipped with an RGB camera, we evaluate the performance of the diffusion policy \citep{chi2023diffusion} both with and without our proposed method. Right: the robot's visual inputs in the task completion.
    }
    \label{fig:real-world-setup}
\end{figure}

\subsection{Real-world Deployment}\label{subsec:real}

\noindent\textbf{Experiment Setup} We deploy a Franka
Emika FR3 robotic arm and a Panda gripper for our real-robot experiments. We focus on the real-world task of picking a colored cup, and placing it onto a plate on the table. We collect a small set of $30$ demonstrations using the Mujoco AR teleoperation interfaces \citep{rayyan2024mujocoar}. We also setup one third-view Realsense D435i camera to provide unposed RGB-only images for observation. We provide an overview of the real-world task setting in~\cref{fig:real-world-setup}. Similar to the experiment setting in \cref{subsec:imitate}, we compare the performance of the state-of-the-art RGB-based policy Diffusion Policy~\cite{chi2023diffusion} with or without our \method representation on task success rate for quantitative analysis.

\begin{table}[t]
\centering
\scriptsize
\caption{\small \textbf{Real-world Experiment Results.} We report the number of successful trials out of all $20$ trials in \textsc{Franka PnP}.}
\resizebox{0.8\linewidth}{!}{
\begin{tabular}{l|c|c}
\toprule
\textsc{Franka-PnP} & \textbf{Diffusion Policy} & \textbf{GWM (Ours)} \\
\midrule
Cup distractor & $6/10$ & \cellcolor{\ourcolor}$\mathbf{7/10}$ \\
Plate distractor & $1/5$ & \cellcolor{\ourcolor}$\mathbf{3/5}$ \\
Table distractor & $0/5$ & \cellcolor{\ourcolor}$\mathbf{3/5}$ \\
\midrule
\textbf{Total} & $7/20$ & \cellcolor{\ourcolor}$\mathbf{13/20}$ \\
\bottomrule
\end{tabular}
}
\label{table:real-world-results}
\end{table}

\noindent\textbf{Results and Analysis} 
As shown in Table~\ref{table:real-world-results}, \method outperforms Diffusion Policy ($\textbf{65}$\% \textit{vs.} $\textbf{35}$\% success rate) on $20$ trials with different initial start positions and object locations (\ie distractors). 
The performance gap widens for novel distractors, demonstrating GWM's superior generalization capabilities. Our approach maintains consistent performance across task variations due to its effective world model that captures task-relevant dynamics while being robust to visual differences. The real-world rollouts are shown in the supplementary file, where the advantage stems primarily from more precise object localization and accurate placement operations. The results demonstrate \method's robust spatial-temporal understanding in real-world robotic manipulation tasks.

\begin{table}[t]
\centering
\scriptsize
\caption{\textbf{Ablation Study} on \texttt{PnP CabToCounter} in \textsc{Robocasa} task suite. We report the reconstruction metrics and the success rates (SR) of imitation learning on the Human-50 dataset.}
\resizebox{\linewidth}{!}{
\begin{tabular}{c|c|cccc|c}
\toprule
GS & 3D VAE & FVD$\downarrow$ & PSNR$\uparrow$ & SSIM$\uparrow$ & LPIPS$\downarrow$ & SR $\uparrow$  \\
\midrule
\no & \no & 67.8 & 27.2 & 88.2 & 5.1 & $4$ \\
\yes & \no & 65.3 & 26.9 & 89.5 & 4.9 & $18$ \\
\yes & \yes & \textbf{62.4} & \textbf{28.1} & \textbf{90.8} & \textbf{4.6} & $\textbf{24}$ \\
\bottomrule
\end{tabular}
}
\label{table:ablation-study}
\end{table}

\subsection{Ablation Analysis}\label{subsec:ablation}
We conduct additional experiments on \textsc{Robocasa} to further verify our design choices.

\noindent\textbf{Choice of Gaussian Splatting} 
As shown in Table~\ref{table:ablation-study}, compared to directly building image-based world model with diffusion transformer on par with \citep{alonso2024diffusion}, introducing Gaussian Splatting significantly improves the success rate (SR) from $4\%$ to $18\%$. While PSNR shows a slight decrease, both SSIM and LPIPS metrics improve, suggesting that Gaussian Splatting provides better 3D consistency across different time steps. This validates our hypothesis that explicit 3D representation enhances spatial understanding for robot learning compared to pure 2D approaches.

\noindent\textbf{Choice of 3D Gaussian VAE}
Further incorporating the 3D VAE component yields consistent improvements across all metrics, including PSNR.
The success rate further improves from $18$\% to $24$\%. The results confirm that our 3D Gaussian VAE efficiently captures the latent structure of the scene, enabling more compact scene representation while maintaining spatial understanding. 


\section{Conclusion}
\label{sec:conclusion}

In this paper, we introduce a novel \ac{gwm} for robotic manipulation that addresses limitations of image-based world models by incorporating robust geometric information. Our approach reconstructs future states by modeling the propagation of Gaussian primitives under robot actions. The method combines a \ac{dit} with a 3D-aware variational autoencoder for precise scene-level future state reconstruction via Gaussian Splatting. We develop a scalable data processing pipeline to facilitate test-time updates within a model-based reinforcement learning framework, extracting aligned Gaussian splats from unposed images. Experiments in both simulated and real-world settings demonstrate the effectiveness of \ac{gwm} in predicting future scenes and training superior policies. 

\section*{Acknowledgement}

This work was supported by Shenzhen Science and Technology Program (JCYJ20240813111903006) and Guangdong Natural Science Funds for Distinguished Young Scholar (No. 2025B1515020012). We would like to thank Ruijie Lu, Junfeng Ni, and Yu Liu from BIGAI General Vision Lab for fruitful discussions.

{
    \small
    \bibliographystyle{ieeenat_fullname}
    \bibliography{main}

\begin{thebibliography}{103}
\providecommand{\natexlab}[1]{#1}
\providecommand{\url}[1]{\texttt{#1}}
\expandafter\ifx\csname urlstyle\endcsname\relax
  \providecommand{\doi}[1]{doi: #1}\else
  \providecommand{\doi}{doi: \begingroup \urlstyle{rm}\Url}\fi

\bibitem[Alonso et~al.(2024)Alonso, Jelley, Micheli, Kanervisto, Storkey, Pearce, and Fleuret]{alonso2024diffusion}
Eloi Alonso, Adam Jelley, Vincent Micheli, Anssi Kanervisto, Amos Storkey, Tim Pearce, and Fran{\c{c}}ois Fleuret.
\newblock Diffusion for world modeling: Visual details matter in atari.
\newblock \emph{arXiv preprint arXiv:2405.12399}, 2024.

\bibitem[Anderson(1982)]{anderson1982reverse}
Brian~DO Anderson.
\newblock Reverse-time diffusion equation models.
\newblock \emph{Stochastic Processes and their Applications}, 12\penalty0 (3):\penalty0 313--326, 1982.

\bibitem[Authors(2024)]{Genesis}
Genesis Authors.
\newblock Genesis: A universal and generative physics engine for robotics and beyond, 2024.

\bibitem[Bansal et~al.(2024)Bansal, Lin, Xie, Zong, Yarom, Bitton, Jiang, Sun, Chang, and Grover]{bansal2024videophy}
Hritik Bansal, Zongyu Lin, Tianyi Xie, Zeshun Zong, Michal Yarom, Yonatan Bitton, Chenfanfu Jiang, Yizhou Sun, Kai-Wei Chang, and Aditya Grover.
\newblock Videophy: Evaluating physical commonsense for video generation.
\newblock \emph{arXiv preprint arXiv:2406.03520}, 2024.

\bibitem[Black et~al.(2024)Black, Brown, Driess, Esmail, Equi, Finn, Fusai, Groom, Hausman, Ichter, et~al.]{black2024pi_0}
Kevin Black, Noah Brown, Danny Driess, Adnan Esmail, Michael Equi, Chelsea Finn, Niccolo Fusai, Lachy Groom, Karol Hausman, Brian Ichter, et~al.
\newblock A vision-language-action flow model for general robot control.
\newblock \emph{arXiv preprint arXiv:2410.24164}, 2024.

\bibitem[Charatan et~al.(2023)Charatan, Li, Tagliasacchi, and Sitzmann]{charatan2023pixelsplat}
David Charatan, Sizhe Li, Andrea Tagliasacchi, and Vincent Sitzmann.
\newblock pixelsplat: 3d gaussian splats from image pairs for scalable generalizable 3d reconstruction.
\newblock \emph{arXiv preprint arXiv:2312.12337}, 2023.

\bibitem[Chen et~al.(2023)Chen, Garcia, Schmid, and Laptev]{chen2023polarnet}
Shizhe Chen, Ricardo Garcia, Cordelia Schmid, and Ivan Laptev.
\newblock Polarnet: 3d point clouds for language-guided robotic manipulation.
\newblock \emph{arXiv preprint arXiv:2309.15596}, 2023.

\bibitem[Chen et~al.(2024)Chen, Ni, Jiang, Zhang, Zhu, and Huang]{chen2023ssr}
Yixin Chen, Junfeng Ni, Nan Jiang, Yaowei Zhang, Yixin Zhu, and Siyuan Huang.
\newblock Single-view 3d scene reconstruction with high-fidelity shape and texture.
\newblock In \emph{Proceedings of International Conference on 3D Vision (3DV)}, 2024.

\bibitem[Chi et~al.(2023)Chi, Xu, Feng, Cousineau, Du, Burchfiel, Tedrake, and Song]{chi2023diffusion}
Cheng Chi, Zhenjia Xu, Siyuan Feng, Eric Cousineau, Yilun Du, Benjamin Burchfiel, Russ Tedrake, and Shuran Song.
\newblock Diffusion policy: Visuomotor policy learning via action diffusion.
\newblock \emph{International Journal of Robotics Research (IJRR)}, page 02783649241273668, 2023.

\bibitem[Dai et~al.(2024)Dai, Wong, Jiang, Wang, Gokmen, Zhang, Wu, and Fei-Fei]{dai2024acdc}
Tianyuan Dai, Josiah Wong, Yunfan Jiang, Chen Wang, Cem Gokmen, Ruohan Zhang, Jiajun Wu, and Li Fei-Fei.
\newblock Acdc: Automated creation of digital cousins for robust policy learning.
\newblock In \emph{8th Annual Conference on Robot Learning}, 2024.

\bibitem[Driess et~al.(2022)Driess, Schubert, Florence, Li, and Toussaint]{driess2022nerfrl}
Danny Driess, Ingmar Schubert, Pete Florence, Yunzhu Li, and Marc Toussaint.
\newblock Reinforcement learning with neural radiance fields.
\newblock \emph{Proceedings of Advances in Neural Information Processing Systems (NeurIPS)}, 2022.

\bibitem[Forrester(1971)]{forrester1971counterintuitive}
Jay~W Forrester.
\newblock Counterintuitive behavior of social systems.
\newblock \emph{Theory and decision}, 2\penalty0 (2):\penalty0 109--140, 1971.

\bibitem[Fu et~al.(2023)Fu, Liu, Kulkarni, Kautz, Efros, and Wang]{fu2023colmap}
Yang Fu, Sifei Liu, Amey Kulkarni, Jan Kautz, Alexei~A Efros, and Xiaolong Wang.
\newblock Colmap-free 3d gaussian splatting.
\newblock \emph{arXiv preprint arXiv:2312.07504}, 2023.

\bibitem[Gao et~al.(2022)Gao, Mu, Shen, Chen, Ren, Chen, Li, Luo, and Lu]{gao2022enhance}
Zeyu Gao, Yao Mu, Ruoyan Shen, Chen Chen, Yangang Ren, Jianyu Chen, Shengbo~Eben Li, Ping Luo, and Yanfeng Lu.
\newblock Enhance sample efficiency and robustness of end-to-end urban autonomous driving via semantic masked world model.
\newblock \emph{arXiv preprint arXiv:2210.04017}, 2022.

\bibitem[Gervet et~al.(2023)Gervet, Xian, Gkanatsios, and Fragkiadaki]{gervet2023act3d}
Theophile Gervet, Zhou Xian, Nikolaos Gkanatsios, and Katerina Fragkiadaki.
\newblock Act3d: 3d feature field transformers for multi-task robotic manipulation.
\newblock In \emph{Conference on Robot Learning (CoRL)}, pages 3949--3965, 2023.

\bibitem[Goyal et~al.(2023)Goyal, Xu, Guo, Blukis, Chao, and Fox]{goyal2023rvt}
Ankit Goyal, Jie Xu, Yijie Guo, Valts Blukis, Yu-Wei Chao, and Dieter Fox.
\newblock Rvt: Robotic view transformer for 3d object manipulation.
\newblock \emph{arXiv preprint arXiv:2306.14896}, 2023.

\bibitem[Goyal et~al.(2024)Goyal, Blukis, Xu, Guo, Chao, and Fox]{goyal2024rvt2}
Ankit Goyal, Valts Blukis, Jie Xu, Yijie Guo, Yu-Wei Chao, and Dieter Fox.
\newblock Rvt-2: Learning precise manipulation from few demonstrations.
\newblock \emph{arXiv preprint arXiv:2406.08545}, 2024.

\bibitem[Ha and Schmidhuber(2018)]{ha2018recurrent}
David Ha and J{\"u}rgen Schmidhuber.
\newblock Recurrent world models facilitate policy evolution.
\newblock \emph{Proceedings of Advances in Neural Information Processing Systems (NeurIPS)}, 31, 2018.

\bibitem[Hafner et~al.(2019)Hafner, Lillicrap, Ba, and Norouzi]{hafner2019dream}
Danijar Hafner, Timothy Lillicrap, Jimmy Ba, and Mohammad Norouzi.
\newblock Dream to control: Learning behaviors by latent imagination.
\newblock \emph{arXiv preprint arXiv:1912.01603}, 2019.

\bibitem[Hafner et~al.(2020)Hafner, Lillicrap, Norouzi, and Ba]{hafner2020mastering}
Danijar Hafner, Timothy Lillicrap, Mohammad Norouzi, and Jimmy Ba.
\newblock Mastering atari with discrete world models.
\newblock \emph{arXiv preprint arXiv:2010.02193}, 2020.

\bibitem[Hafner et~al.(2022)Hafner, Lee, Fischer, and Abbeel]{hafner2022deep}
Danijar Hafner, Kuang-Huei Lee, Ian Fischer, and Pieter Abbeel.
\newblock Deep hierarchical planning from pixels.
\newblock \emph{Proceedings of Advances in Neural Information Processing Systems (NeurIPS)}, 35:\penalty0 26091--26104, 2022.

\bibitem[Hafner et~al.(2023)Hafner, Pasukonis, Ba, and Lillicrap]{hafner2023mastering}
Danijar Hafner, Jurgis Pasukonis, Jimmy Ba, and Timothy Lillicrap.
\newblock Mastering diverse domains through world models.
\newblock \emph{arXiv preprint arXiv:2301.04104}, 2023.

\bibitem[Hansen et~al.(2023)Hansen, Su, and Wang]{hansen2023td}
Nicklas Hansen, Hao Su, and Xiaolong Wang.
\newblock Td-mpc2: Scalable, robust world models for continuous control.
\newblock \emph{arXiv preprint arXiv:2310.16828}, 2023.

\bibitem[Ho et~al.(2020{\natexlab{a}})Ho, Jain, and Abbeel]{ho2020DDPM}
Jonathan Ho, Ajay Jain, and Pieter Abbeel.
\newblock Denoising diffusion probabilistic models.
\newblock \emph{Advances in Neural Information Processing Systems}, 33:\penalty0 6840--6851, 2020{\natexlab{a}}.

\bibitem[Ho et~al.(2020{\natexlab{b}})Ho, Jain, and Abbeel]{ho2020denoising}
Jonathan Ho, Ajay Jain, and Pieter Abbeel.
\newblock Denoising diffusion probabilistic models.
\newblock \emph{Proceedings of Advances in Neural Information Processing Systems (NeurIPS)}, 2020{\natexlab{b}}.

\bibitem[Hu et~al.(2022)Hu, Corrado, Griffiths, Murez, Gurau, Yeo, Kendall, Cipolla, and Shotton]{hu2022model}
Anthony Hu, Gianluca Corrado, Nicolas Griffiths, Zachary Murez, Corina Gurau, Hudson Yeo, Alex Kendall, Roberto Cipolla, and Jamie Shotton.
\newblock Model-based imitation learning for urban driving.
\newblock \emph{Proceedings of Advances in Neural Information Processing Systems (NeurIPS)}, 35:\penalty0 20703--20716, 2022.

\bibitem[Hu et~al.(2023)Hu, Russell, Yeo, Murez, Fedoseev, Kendall, Shotton, and Corrado]{hu2023gaia}
Anthony Hu, Lloyd Russell, Hudson Yeo, Zak Murez, George Fedoseev, Alex Kendall, Jamie Shotton, and Gianluca Corrado.
\newblock Gaia-1: A generative world model for autonomous driving.
\newblock \emph{arXiv preprint arXiv:2309.17080}, 2023.

\bibitem[Huang et~al.(2024)Huang, Sun, Yang, Lyu, Cao, and Qi]{huang2024sc}
Yi-Hua Huang, Yang-Tian Sun, Ziyi Yang, Xiaoyang Lyu, Yan-Pei Cao, and Xiaojuan Qi.
\newblock Sc-gs: Sparse-controlled gaussian splatting for editable dynamic scenes.
\newblock In \emph{Proceedings of Conference on Computer Vision and Pattern Recognition (CVPR)}, 2024.

\bibitem[Huynh-Thu and Ghanbari(2008)]{psnr}
Quan Huynh-Thu and Mohammed Ghanbari.
\newblock Scope of validity of psnr in image/video quality assessment.
\newblock \emph{Electronics letters}, 44\penalty0 (13):\penalty0 800--801, 2008.

\bibitem[Hyv{\"a}rinen and Dayan(2005)]{hyvarinen2005estimation}
Aapo Hyv{\"a}rinen and Peter Dayan.
\newblock Estimation of non-normalized statistical models by score matching.
\newblock \emph{Journal of Machine Learning Research (JMLR)}, 6\penalty0 (4), 2005.

\bibitem[Janner et~al.(2019)Janner, Fu, Zhang, and Levine]{janner2021trust}
Michael Janner, Justin Fu, Marvin Zhang, and Sergey Levine.
\newblock When to trust your model: Model-based policy optimization.
\newblock In \emph{Proceedings of Advances in Neural Information Processing Systems (NeurIPS)}, 2019.

\bibitem[Jiang et~al.(2021)Jiang, Zhu, Svetlik, Fang, and Zhu]{jiang2021giga}
Zhenyu Jiang, Yifeng Zhu, Maxwell Svetlik, Kuan Fang, and Yuke Zhu.
\newblock Synergies between affordance and geometry: 6-dof grasp detection via implicit representations.
\newblock \emph{arXiv preprint arXiv:2104.01542}, 2021.

\bibitem[Karras et~al.(2022)Karras, Aittala, Aila, and Laine]{karras2022elucidating}
Tero Karras, Miika Aittala, Timo Aila, and Samuli Laine.
\newblock Elucidating the design space of diffusion-based generative models.
\newblock \emph{Advances in Neural Information Processing Systems}, 35:\penalty0 26565--26577, 2022.

\bibitem[Keetha et~al.(2024)Keetha, Karhade, Jatavallabhula, Yang, Scherer, Ramanan, and Luiten]{keetha2024splatam}
Nikhil Keetha, Jay Karhade, Krishna~Murthy Jatavallabhula, Gengshan Yang, Sebastian Scherer, Deva Ramanan, and Jonathon Luiten.
\newblock Splatam: Splat track \& map 3d gaussians for dense rgb-d slam.
\newblock In \emph{Proceedings of the IEEE/CVF Conference on Computer Vision and Pattern Recognition}, pages 21357--21366, 2024.

\bibitem[Kerbl et~al.(2023)Kerbl, Kopanas, Leimk{\"u}hler, and Drettakis]{kerbl20233d}
Bernhard Kerbl, Georgios Kopanas, Thomas Leimk{\"u}hler, and George Drettakis.
\newblock 3d gaussian splatting for real-time radiance field rendering.
\newblock \emph{TOG}, 42\penalty0 (4), 2023.

\bibitem[Kim et~al.(2024)Kim, Pertsch, Karamcheti, Xiao, Balakrishna, Nair, Rafailov, Foster, Lam, Sanketi, et~al.]{kim2024openvla}
Moo~Jin Kim, Karl Pertsch, Siddharth Karamcheti, Ted Xiao, Ashwin Balakrishna, Suraj Nair, Rafael Rafailov, Ethan Foster, Grace Lam, Pannag Sanketi, et~al.
\newblock Openvla: An open-source vision-language-action model.
\newblock \emph{arXiv preprint arXiv:2406.09246}, 2024.

\bibitem[Leroy et~al.(2024)Leroy, Cabon, and Revaud]{leroy2024grounding}
Vincent Leroy, Yohann Cabon, and J{\'e}r{\^o}me Revaud.
\newblock Grounding image matching in 3d with mast3r.
\newblock \emph{arXiv preprint arXiv:2406.09756}, 2024.

\bibitem[Li et~al.(2024{\natexlab{a}})Li, Liu, Li, Han, Geng, Wang, Zhu, Zhu, and Huang]{li2024ag2manip}
Puhao Li, Tengyu Liu, Yuyang Li, Muzhi Han, Haoran Geng, Shu Wang, Yixin Zhu, Song-Chun Zhu, and Siyuan Huang.
\newblock Ag2manip: Learning novel manipulation skills with agent-agnostic visual and action representations.
\newblock In \emph{Proceedings of International Conference on Intelligent Robots and Systems (IROS)}, pages 573--580. IEEE, 2024{\natexlab{a}}.

\bibitem[Li et~al.(2025)Li, Wu, Xi, Li, Huang, Zhang, Chen, Wang, Zhu, Liu, et~al.]{li2025controlvla}
Puhao Li, Yingying Wu, Ziheng Xi, Wanlin Li, Yuzhe Huang, Zhiyuan Zhang, Yinghan Chen, Jianan Wang, Song-Chun Zhu, Tengyu Liu, et~al.
\newblock Controlvla: Few-shot object-centric adaptation for pre-trained vision-language-action models.
\newblock \emph{arXiv preprint arXiv:2506.16211}, 2025.

\bibitem[Li et~al.(2024{\natexlab{b}})Li, Hsu, Gu, Pertsch, Mees, Walke, Fu, Lunawat, Sieh, Kirmani, et~al.]{li2024evaluating}
Xuanlin Li, Kyle Hsu, Jiayuan Gu, Karl Pertsch, Oier Mees, Homer~Rich Walke, Chuyuan Fu, Ishikaa Lunawat, Isabel Sieh, Sean Kirmani, et~al.
\newblock Evaluating real-world robot manipulation policies in simulation.
\newblock \emph{arXiv preprint arXiv:2405.05941}, 2024{\natexlab{b}}.

\bibitem[Li et~al.(2022)Li, Li, Sitzmann, Agrawal, and Torralba]{li20223d}
Yunzhu Li, Shuang Li, Vincent Sitzmann, Pulkit Agrawal, and Antonio Torralba.
\newblock 3d neural scene representations for visuomotor control.
\newblock In \emph{Conference on Robot Learning (CoRL)}, pages 112--123, 2022.

\bibitem[Lim et~al.(2022)Lim, Huang, Chen, Wang, Ichnowski, Seita, Laskey, and Goldberg]{lim2022real2sim2real}
Vincent Lim, Huang Huang, Lawrence~Yunliang Chen, Jonathan Wang, Jeffrey Ichnowski, Daniel Seita, Michael Laskey, and Ken Goldberg.
\newblock Real2sim2real: Self-supervised learning of physical single-step dynamic actions for planar robot casting.
\newblock In \emph{2022 International Conference on Robotics and Automation (ICRA)}, pages 8282--8289. IEEE, 2022.

\bibitem[Lin et~al.(2023)Lin, Florence, Zeng, Barron, Du, Ma, Simeonov, Garcia, and Isola]{lin2023mira}
Yen-Chen Lin, Pete Florence, Andy Zeng, Jonathan~T Barron, Yilun Du, Wei-Chiu Ma, Anthony Simeonov, Alberto~Rodriguez Garcia, and Phillip Isola.
\newblock Mira: Mental imagery for robotic affordances.
\newblock In \emph{Conference on Robot Learning (CoRL)}, pages 1916--1927, 2023.

\bibitem[Liu et~al.(2024{\natexlab{a}})Liu, Arthur, He, Seita, and Sukhatme]{liu2024voxact}
I Liu, Chun Arthur, Sicheng He, Daniel Seita, and Gaurav Sukhatme.
\newblock Voxact-b: Voxel-based acting and stabilizing policy for bimanual manipulation.
\newblock \emph{arXiv preprint arXiv:2407.04152}, 2024{\natexlab{a}}.

\bibitem[Liu et~al.(2024{\natexlab{b}})Liu, Wu, Li, Tan, Chen, Wang, Xu, Su, and Zhu]{liu2024rdt}
Songming Liu, Lingxuan Wu, Bangguo Li, Hengkai Tan, Huayu Chen, Zhengyi Wang, Ke Xu, Hang Su, and Jun Zhu.
\newblock Rdt-1b: a diffusion foundation model for bimanual manipulation.
\newblock \emph{arXiv preprint arXiv:2410.07864}, 2024{\natexlab{b}}.

\bibitem[Liu et~al.(2024{\natexlab{c}})Liu, Li, Yang, and Yuan]{liu2024endogaussian}
Yifan Liu, Chenxin Li, Chen Yang, and Yixuan Yuan.
\newblock Endogaussian: Gaussian splatting for deformable surgical scene reconstruction.
\newblock \emph{arXiv preprint arXiv:2401.12561}, 2024{\natexlab{c}}.

\bibitem[Liu et~al.(2025)Liu, Jia, Lu, Ni, Zhu, and Huang]{liu2025building}
Yu Liu, Baoxiong Jia, Ruijie Lu, Junfeng Ni, Song-Chun Zhu, and Siyuan Huang.
\newblock Building interactable replicas of complex articulated objects via gaussian splatting.
\newblock In \emph{Proceedings of International Conference on Learning Representations (ICLR)}, 2025.

\bibitem[Lou et~al.(2024)Lou, Liu, Pan, Geng, Chen, Ma, Li, Wang, Feng, Shi, et~al.]{lou2024robo}
Haozhe Lou, Yurong Liu, Yike Pan, Yiran Geng, Jianteng Chen, Wenlong Ma, Chenglong Li, Lin Wang, Hengzhen Feng, Lu Shi, et~al.
\newblock Robo-gs: A physics consistent spatial-temporal model for robotic arm with hybrid representation.
\newblock \emph{arXiv preprint arXiv:2408.14873}, 2024.

\bibitem[Lu et~al.(2024)Lu, Zhang, Wang, Liu, Lu, and Tang]{lu2024manigaussian}
Guanxing Lu, Shiyi Zhang, Ziwei Wang, Changliu Liu, Jiwen Lu, and Yansong Tang.
\newblock Manigaussian: Dynamic gaussian splatting for multi-task robotic manipulation.
\newblock In \emph{European Conference on Computer Vision}, pages 349--366. Springer, 2024.

\bibitem[Lu et~al.(2025{\natexlab{a}})Lu, Chen, Liu, Tang, Ni, Wan, Zeng, and Huang]{lu2025taco}
Ruijie Lu, Yixin Chen, Yu Liu, Jiaxiang Tang, Junfeng Ni, Diwen Wan, Gang Zeng, and Siyuan Huang.
\newblock Taco: Taming diffusion for in-the-wild video amodal completion.
\newblock \emph{arXiv preprint arXiv:2503.12049}, 2025{\natexlab{a}}.

\bibitem[Lu et~al.(2025{\natexlab{b}})Lu, Liu, Tang, Ni, Wang, Wan, Zeng, Chen, and Huang]{lu2025dreamart}
Ruijie Lu, Yu Liu, Jiaxiang Tang, Junfeng Ni, Yuxiang Wang, Diwen Wan, Gang Zeng, Yixin Chen, and Siyuan Huang.
\newblock Dreamart: Generating interactable articulated objects from a single image.
\newblock \emph{arXiv preprint arXiv:2507.05763}, 2025{\natexlab{b}}.

\bibitem[Luiten et~al.(2024)Luiten, Kopanas, Leibe, and Ramanan]{luiten2024dynamic}
Jonathon Luiten, Georgios Kopanas, Bastian Leibe, and Deva Ramanan.
\newblock Dynamic 3d gaussians: Tracking by persistent dynamic view synthesis.
\newblock In \emph{Proceedings of International Conference on 3D Vision (3DV)}, 2024.

\bibitem[Luo et~al.(2024{\natexlab{a}})Luo, Hu, Xu, Tan, Berg, Sharma, Schaal, Finn, Gupta, and Levine]{luo2024serl}
Jianlan Luo, Zheyuan Hu, Charles Xu, You~Liang Tan, Jacob Berg, Archit Sharma, Stefan Schaal, Chelsea Finn, Abhishek Gupta, and Sergey Levine.
\newblock Serl: A software suite for sample-efficient robotic reinforcement learning.
\newblock In \emph{Proceedings of International Conference on Robotics and Automation (ICRA)}, pages 16961--16969. IEEE, 2024{\natexlab{a}}.

\bibitem[Luo et~al.(2024{\natexlab{b}})Luo, Xu, Wu, and Levine]{luo2024precise}
Jianlan Luo, Charles Xu, Jeffrey Wu, and Sergey Levine.
\newblock Precise and dexterous robotic manipulation via human-in-the-loop reinforcement learning.
\newblock \emph{arXiv preprint arXiv:2410.21845}, 2024{\natexlab{b}}.

\bibitem[Mandlekar et~al.(2023)Mandlekar, Nasiriany, Wen, Akinola, Narang, Fan, Zhu, and Fox]{mandlekar2023mimicgen}
Ajay Mandlekar, Soroush Nasiriany, Bowen Wen, Iretiayo Akinola, Yashraj Narang, Linxi Fan, Yuke Zhu, and Dieter Fox.
\newblock Mimicgen: A data generation system for scalable robot learning using human demonstrations.
\newblock \emph{arXiv preprint arXiv:2310.17596}, 2023.

\bibitem[Micheli et~al.(2023)Micheli, Alonso, and Fleuret]{iris2023}
Vincent Micheli, Eloi Alonso, and Fran{\c{c}}ois Fleuret.
\newblock Transformers are sample-efficient world models.
\newblock \emph{International Conference on Learning Representations}, 2023.

\bibitem[Mildenhall et~al.(2021)Mildenhall, Srinivasan, Tancik, Barron, Ramamoorthi, and Ng]{mildenhall2021nerf}
Ben Mildenhall, Pratul~P Srinivasan, Matthew Tancik, Jonathan~T Barron, Ravi Ramamoorthi, and Ren Ng.
\newblock Nerf: Representing scenes as neural radiance fields for view synthesis.
\newblock \emph{CACM}, 65\penalty0 (1):\penalty0 99--106, 2021.

\bibitem[Mu et~al.(2025)Mu, Chen, Chen, Peng, Lan, Gao, Liang, Yu, Zou, Xu, et~al.]{mu2024robotwin}
Yao Mu, Tianxing Chen, Zanxin Chen, Shijia Peng, Zhiqian Lan, Zeyu Gao, Zhixuan Liang, Qiaojun Yu, Yude Zou, Mingkun Xu, et~al.
\newblock Robotwin: Dual-arm robot benchmark with generative digital twins.
\newblock In \emph{Proceedings of Conference on Computer Vision and Pattern Recognition (CVPR)}, pages 27649--27660, 2025.

\bibitem[Nasiriany et~al.(2024)Nasiriany, Maddukuri, Zhang, Parikh, Lo, Joshi, Mandlekar, and Zhu]{nasiriany2024robocasa}
Soroush Nasiriany, Abhiram Maddukuri, Lance Zhang, Adeet Parikh, Aaron Lo, Abhishek Joshi, Ajay Mandlekar, and Yuke Zhu.
\newblock Robocasa: Large-scale simulation of everyday tasks for generalist robots.
\newblock \emph{arXiv preprint arXiv:2406.02523}, 2024.

\bibitem[Peebles and Xie(2023)]{peebles2023scalable}
William Peebles and Saining Xie.
\newblock Scalable diffusion models with transformers.
\newblock In \emph{Proceedings of International Conference on Computer Vision (ICCV)}, pages 4195--4205, 2023.

\bibitem[Perez et~al.(2018)Perez, Strub, De~Vries, Dumoulin, and Courville]{perez2018film}
Ethan Perez, Florian Strub, Harm De~Vries, Vincent Dumoulin, and Aaron Courville.
\newblock Film: Visual reasoning with a general conditioning layer.
\newblock In \emph{Proceedings of AAAI Conference on Artificial Intelligence (AAAI)}, 2018.

\bibitem[Peri et~al.(2024)Peri, Lee, Kim, Fuxin, Hermans, and Lee]{peri2024point}
Skand Peri, Iain Lee, Chanho Kim, Li Fuxin, Tucker Hermans, and Stefan Lee.
\newblock Point cloud models improve visual robustness in robotic learners.
\newblock In \emph{Proceedings of International Conference on Robotics and Automation (ICRA)}, 2024.

\bibitem[Qureshi et~al.(2024)Qureshi, Garg, Yandun, Held, Kantor, and Silwal]{qureshi2024splatsim}
Mohammad~Nomaan Qureshi, Sparsh Garg, Francisco Yandun, David Held, George Kantor, and Abhishesh Silwal.
\newblock Splatsim: Zero-shot sim2real transfer of rgb manipulation policies using gaussian splatting.
\newblock \emph{arXiv preprint arXiv:2409.10161}, 2024.

\bibitem[Radford et~al.(2018)Radford, Narasimhan, Salimans, Sutskever, et~al.]{radford2018improving}
Alec Radford, Karthik Narasimhan, Tim Salimans, Ilya Sutskever, et~al.
\newblock Improving language understanding by generative pre-training.
\newblock 2018.

\bibitem[Rayyan(2024)]{rayyan2024mujocoar}
Omar Rayyan.
\newblock {MuJoCoAR}: Phone teleoperation for robots, 2024.

\bibitem[Robine et~al.(2023)Robine, H{\"o}ftmann, Uelwer, and Harmeling]{robine2023transformer}
Jan Robine, Marc H{\"o}ftmann, Tobias Uelwer, and Stefan Harmeling.
\newblock Transformer-based world models are happy with 100k interactions.
\newblock \emph{International Conference on Learning Representations}, 2023.

\bibitem[Schrittwieser et~al.(2020)Schrittwieser, Antonoglou, Hubert, Simonyan, Sifre, Schmitt, Guez, Lockhart, Hassabis, Graepel, et~al.]{schrittwieser2020mastering}
Julian Schrittwieser, Ioannis Antonoglou, Thomas Hubert, Karen Simonyan, Laurent Sifre, Simon Schmitt, Arthur Guez, Edward Lockhart, Demis Hassabis, Thore Graepel, et~al.
\newblock Mastering atari, go, chess and shogi by planning with a learned model.
\newblock \emph{Nature}, 588\penalty0 (7839):\penalty0 604--609, 2020.

\bibitem[Seo et~al.(2023)Seo, Hafner, Liu, Liu, James, Lee, and Abbeel]{seo2023masked}
Younggyo Seo, Danijar Hafner, Hao Liu, Fangchen Liu, Stephen James, Kimin Lee, and Pieter Abbeel.
\newblock Masked world models for visual control.
\newblock In \emph{Conference on Robot Learning (CoRL)}, pages 1332--1344, 2023.

\bibitem[Shim et~al.(2023)Shim, Lee, and Kim]{shim2023snerl}
Dongseok Shim, Seungjae Lee, and H~Jin Kim.
\newblock Snerl: Semantic-aware neural radiance fields for reinforcement learning.
\newblock In \emph{Proceedings of International Conference on Machine Learning (ICML)}, 2023.

\bibitem[Shridhar et~al.(2023)Shridhar, Manuelli, and Fox]{shridhar2023peract}
Mohit Shridhar, Lucas Manuelli, and Dieter Fox.
\newblock Perceiver-actor: A multi-task transformer for robotic manipulation.
\newblock In \emph{Conference on Robot Learning (CoRL)}, 2023.

\bibitem[Smart et~al.(2024)Smart, Zheng, Laina, and Prisacariu]{smart2024splatt3r}
Brandon Smart, Chuanxia Zheng, Iro Laina, and Victor~Adrian Prisacariu.
\newblock Splatt3r: Zero-shot gaussian splatting from uncalibarated image pairs.
\newblock \emph{arXiv preprint arXiv:2408.13912}, 2024.

\bibitem[Sohl-Dickstein et~al.(2015)Sohl-Dickstein, Weiss, Maheswaranathan, and Ganguli]{sohl2015difforigin}
Jascha Sohl-Dickstein, Eric Weiss, Niru Maheswaranathan, and Surya Ganguli.
\newblock Deep unsupervised learning using nonequilibrium thermodynamics.
\newblock \emph{International Conference on Machine Learning}, 2015.

\bibitem[Song et~al.(2020)Song, Sohl-Dickstein, Kingma, Kumar, Ermon, and Poole]{song_sde}
Yang Song, Jascha Sohl-Dickstein, Diederik~P Kingma, Abhishek Kumar, Stefano Ermon, and Ben Poole.
\newblock Score-based generative modeling through stochastic differential equations.
\newblock \emph{International Conference on Learning Representations}, 2020.

\bibitem[Su et~al.(2024)Su, Ahmed, Lu, Pan, Bo, and Liu]{su2024roformer}
Jianlin Su, Murtadha Ahmed, Yu Lu, Shengfeng Pan, Wen Bo, and Yunfeng Liu.
\newblock Roformer: Enhanced transformer with rotary position embedding.
\newblock \emph{Neurocomputing}, 568:\penalty0 127063, 2024.

\bibitem[Szymanowicz et~al.(2023)Szymanowicz, Rupprecht, and Vedaldi]{szymanowicz2023splatter}
Stanislaw Szymanowicz, Christian Rupprecht, and Andrea Vedaldi.
\newblock Splatter image: Ultra-fast single-view 3d reconstruction.
\newblock \emph{arXiv preprint arXiv:2312.13150}, 2023.

\bibitem[Team et~al.(2024)Team, Ghosh, Walke, Pertsch, Black, Mees, Dasari, Hejna, Kreiman, Xu, et~al.]{team2024octo}
Octo~Model Team, Dibya Ghosh, Homer Walke, Karl Pertsch, Kevin Black, Oier Mees, Sudeep Dasari, Joey Hejna, Tobias Kreiman, Charles Xu, et~al.
\newblock Octo: An open-source generalist robot policy.
\newblock \emph{arXiv preprint arXiv:2405.12213}, 2024.

\bibitem[Unterthiner et~al.(2018)Unterthiner, Van~Steenkiste, Kurach, Marinier, Michalski, and Gelly]{fvd}
Thomas Unterthiner, Sjoerd Van~Steenkiste, Karol Kurach, Raphael Marinier, Marcin Michalski, and Sylvain Gelly.
\newblock Towards accurate generative models of video: A new metric \& challenges.
\newblock \emph{arXiv preprint arXiv:1812.01717}, 2018.

\bibitem[Vaswani et~al.(2017)Vaswani, Shazeer, Parmar, Uszkoreit, Jones, Gomez, Kaiser, and Polosukhin]{vaswani2017attention}
Ashish Vaswani, Noam Shazeer, Niki Parmar, Jakob Uszkoreit, Llion Jones, Aidan~N Gomez, {\L}ukasz Kaiser, and Illia Polosukhin.
\newblock Attention is all you need.
\newblock In \emph{Proceedings of Advances in Neural Information Processing Systems (NeurIPS)}, 2017.

\bibitem[Wang et~al.(2023{\natexlab{a}})Wang, Ling, Yuan, Shridhar, Bao, Qin, Wang, Xu, and Wang]{wang2023gensim}
Lirui Wang, Yiyang Ling, Zhecheng Yuan, Mohit Shridhar, Chen Bao, Yuzhe Qin, Bailin Wang, Huazhe Xu, and Xiaolong Wang.
\newblock Gensim: Generating robotic simulation tasks via large language models.
\newblock \emph{arXiv preprint arXiv:2310.01361}, 2023{\natexlab{a}}.

\bibitem[Wang et~al.(2024)Wang, Leroy, Cabon, Chidlovskii, and Revaud]{wang2024dust3r}
Shuzhe Wang, Vincent Leroy, Yohann Cabon, Boris Chidlovskii, and Jerome Revaud.
\newblock Dust3r: Geometric 3d vision made easy.
\newblock In \emph{Proceedings of the IEEE/CVF Conference on Computer Vision and Pattern Recognition}, pages 20697--20709, 2024.

\bibitem[Wang et~al.(2023{\natexlab{b}})Wang, Zhu, Huang, Chen, and Lu]{wang2023drivedreamer}
Xiaofeng Wang, Zheng Zhu, Guan Huang, Xinze Chen, and Jiwen Lu.
\newblock Drivedreamer: Towards real-world-driven world models for autonomous driving.
\newblock \emph{arXiv preprint arXiv:2309.09777}, 2023{\natexlab{b}}.

\bibitem[Wang et~al.(2004)Wang, Bovik, Sheikh, and Simoncelli]{ssim}
Zhou Wang, Alan~C Bovik, Hamid~R Sheikh, and Eero~P Simoncelli.
\newblock Image quality assessment: from error visibility to structural similarity.
\newblock \emph{Proceedings of Transactions on Image Processing (TIP)}, 13\penalty0 (4):\penalty0 600--612, 2004.

\bibitem[Wu et~al.(2025)Wu, Yin, Feng, He, Li, Hao, and Long]{wu2025ivideogpt}
Jialong Wu, Shaofeng Yin, Ningya Feng, Xu He, Dong Li, Jianye Hao, and Mingsheng Long.
\newblock ivideogpt: Interactive videogpts are scalable world models.
\newblock \emph{Proceedings of Advances in Neural Information Processing Systems (NeurIPS)}, 37:\penalty0 68082--68119, 2025.

\bibitem[Wu et~al.(2023)Wu, Escontrela, Hafner, Abbeel, and Goldberg]{wu2023daydreamer}
Philipp Wu, Alejandro Escontrela, Danijar Hafner, Pieter Abbeel, and Ken Goldberg.
\newblock Daydreamer: World models for physical robot learning.
\newblock In \emph{Conference on Robot Learning (CoRL)}, pages 2226--2240, 2023.

\bibitem[Xie et~al.(2024)Xie, Zong, Qiu, Li, Feng, Yang, and Jiang]{xie2024physgaussian}
Tianyi Xie, Zeshun Zong, Yuxing Qiu, Xuan Li, Yutao Feng, Yin Yang, and Chenfanfu Jiang.
\newblock Physgaussian: Physics-integrated 3d gaussians for generative dynamics.
\newblock In \emph{Proceedings of Conference on Computer Vision and Pattern Recognition (CVPR)}, 2024.

\bibitem[Xu et~al.(2024{\natexlab{a}})Xu, Yuan, Mardani, Liu, Song, Wang, and Vahdat]{xu2024agg}
Dejia Xu, Ye Yuan, Morteza Mardani, Sifei Liu, Jiaming Song, Zhangyang Wang, and Arash Vahdat.
\newblock Agg: Amortized generative 3d gaussians for single image to 3d.
\newblock \emph{arXiv preprint arXiv:2401.04099}, 2024{\natexlab{a}}.

\bibitem[Xu et~al.(2024{\natexlab{b}})Xu, Shi, Yifan, Chen, Yang, Peng, Shen, and Wetzstein]{xu2024grm}
Yinghao Xu, Zifan Shi, Wang Yifan, Hansheng Chen, Ceyuan Yang, Sida Peng, Yujun Shen, and Gordon Wetzstein.
\newblock Grm: Large gaussian reconstruction model for efficient 3d reconstruction and generation.
\newblock \emph{arXiv preprint arXiv:2403.14621}, 2024{\natexlab{b}}.

\bibitem[Yang et~al.(2023)Yang, Du, Ghasemipour, Tompson, Schuurmans, and Abbeel]{yang2023learning}
Mengjiao Yang, Yilun Du, Kamyar Ghasemipour, Jonathan Tompson, Dale Schuurmans, and Pieter Abbeel.
\newblock Learning interactive real-world simulators.
\newblock \emph{arXiv preprint arXiv:2310.06114}, 2023.

\bibitem[Yarats et~al.(2022)Yarats, Fergus, Lazaric, and Pinto]{yarats2021mastering}
Denis Yarats, Rob Fergus, Alessandro Lazaric, and Lerrel Pinto.
\newblock Mastering visual continuous control: Improved data-augmented reinforcement learning.
\newblock In \emph{Proceedings of International Conference on Learning Representations (ICLR)}, 2022.

\bibitem[Ye et~al.(2021)Ye, Liu, Kurutach, Abbeel, and Gao]{ye2021mastering}
Weirui Ye, Shaohuai Liu, Thanard Kurutach, Pieter Abbeel, and Yang Gao.
\newblock Mastering atari games with limited data.
\newblock \emph{Proceedings of Advances in Neural Information Processing Systems (NeurIPS)}, 34:\penalty0 25476--25488, 2021.

\bibitem[Yu et~al.(2020)Yu, Quillen, He, Julian, Hausman, Finn, and Levine]{yu2020meta}
Tianhe Yu, Deirdre Quillen, Zhanpeng He, Ryan Julian, Karol Hausman, Chelsea Finn, and Sergey Levine.
\newblock Meta-world: A benchmark and evaluation for multi-task and meta reinforcement learning.
\newblock In \emph{Conference on Robot Learning (CoRL)}, pages 1094--1100, 2020.

\bibitem[Ze et~al.(2023)Ze, Yan, Wu, Macaluso, Ge, Ye, Hansen, Li, and Wang]{ze2023gnfactor}
Yanjie Ze, Ge Yan, Yueh-Hua Wu, Annabella Macaluso, Yuying Ge, Jianglong Ye, Nicklas Hansen, Li~Erran Li, and Xiaolong Wang.
\newblock Gnfactor: Multi-task real robot learning with generalizable neural feature fields.
\newblock In \emph{Conference on Robot Learning (CoRL)}, pages 284--301. PMLR, 2023.

\bibitem[Zhang and Sennrich(2019)]{zhang2019root}
Biao Zhang and Rico Sennrich.
\newblock Root mean square layer normalization.
\newblock \emph{Proceedings of Advances in Neural Information Processing Systems (NeurIPS)}, 32, 2019.

\bibitem[Zhang et~al.(2023{\natexlab{a}})Zhang, Tang, Niessner, and Wonka]{zhang20233dshape2vecset}
Biao Zhang, Jiapeng Tang, Matthias Niessner, and Peter Wonka.
\newblock 3dshape2vecset: A 3d shape representation for neural fields and generative diffusion models.
\newblock \emph{ACM Transactions On Graphics (TOG)}, 42\penalty0 (4):\penalty0 1--16, 2023{\natexlab{a}}.

\bibitem[Zhang et~al.(2024)Zhang, Cheng, Yang, Wang, Zhao, Tang, Chen, and Guo]{zhang2024gaussiancube}
Bowen Zhang, Yiji Cheng, Jiaolong Yang, Chunyu Wang, Feng Zhao, Yansong Tang, Dong Chen, and Baining Guo.
\newblock Gaussiancube: Structuring gaussian splatting using optimal transport for 3d generative modeling.
\newblock \emph{arXiv preprint arXiv:2403.19655}, 2024.

\bibitem[Zhang et~al.(2018)Zhang, Isola, Efros, Shechtman, and Wang]{lpips}
Richard Zhang, Phillip Isola, Alexei~A Efros, Eli Shechtman, and Oliver Wang.
\newblock The unreasonable effectiveness of deep features as a perceptual metric.
\newblock In \emph{Proceedings of Conference on Computer Vision and Pattern Recognition (CVPR)}, 2018.

\bibitem[Zhang et~al.(2023{\natexlab{b}})Zhang, Wang, Sun, Yuan, and Huang]{zhang2023storm}
Weipu Zhang, Gang Wang, Jian Sun, Yetian Yuan, and Gao Huang.
\newblock Storm: Efficient stochastic transformer based world models for reinforcement learning.
\newblock In \emph{Thirty-seventh Conference on Neural Information Processing Systems}, 2023{\natexlab{b}}.

\bibitem[Zheng et~al.(2023{\natexlab{a}})Zheng, Zhou, Shao, Liu, Zhang, Nie, and Liu]{zheng2023gps}
Shunyuan Zheng, Boyao Zhou, Ruizhi Shao, Boning Liu, Shengping Zhang, Liqiang Nie, and Yebin Liu.
\newblock Gps-gaussian: Generalizable pixel-wise 3d gaussian splatting for real-time human novel view synthesis.
\newblock \emph{arXiv preprint arXiv:2312.02155}, 2023{\natexlab{a}}.

\bibitem[Zheng et~al.(2023{\natexlab{b}})Zheng, Chen, Huang, Zhang, Duan, and Lu]{zheng2023occworld}
Wenzhao Zheng, Weiliang Chen, Yuanhui Huang, Borui Zhang, Yueqi Duan, and Jiwen Lu.
\newblock Occworld: Learning a 3d occupancy world model for autonomous driving.
\newblock \emph{arXiv preprint arXiv:2311.16038}, 2023{\natexlab{b}}.

\bibitem[Zhou et~al.(2024)Zhou, Lin, Shan, Wang, Sun, and Yang]{zhou2024drivinggaussian}
Xiaoyu Zhou, Zhiwei Lin, Xiaojun Shan, Yongtao Wang, Deqing Sun, and Ming-Hsuan Yang.
\newblock Drivinggaussian: Composite gaussian splatting for surrounding dynamic autonomous driving scenes.
\newblock In \emph{Proceedings of the IEEE/CVF Conference on Computer Vision and Pattern Recognition}, pages 21634--21643, 2024.

\bibitem[Zhu et~al.(2025)Zhu, Wang, Huang, Ye, Ouyang, and He]{zhu2025point}
Haoyi Zhu, Yating Wang, Di Huang, Weicai Ye, Wanli Ouyang, and Tong He.
\newblock Point cloud matters: Rethinking the impact of different observation spaces on robot learning.
\newblock \emph{Proceedings of Advances in Neural Information Processing Systems (NeurIPS)}, 2025.

\bibitem[Zou et~al.(2023)Zou, Yu, Guo, Li, Liang, Cao, and Zhang]{zou2023triplane}
Zi-Xin Zou, Zhipeng Yu, Yuan-Chen Guo, Yangguang Li, Ding Liang, Yan-Pei Cao, and Song-Hai Zhang.
\newblock Triplane meets gaussian splatting: Fast and generalizable single-view 3d reconstruction with transformers.
\newblock \emph{arXiv preprint arXiv:2312.09147}, 2023.

\bibitem[Zuo et~al.(2025)Zuo, Zheng, Huang, Zhou, and Lu]{zuo2025gaussianworld}
Sicheng Zuo, Wenzhao Zheng, Yuanhui Huang, Jie Zhou, and Jiwen Lu.
\newblock Gaussianworld: Gaussian world model for streaming 3d occupancy prediction.
\newblock In \emph{Proceedings of Conference on Computer Vision and Pattern Recognition (CVPR)}, pages 6772--6781, 2025.

\end{thebibliography}
}

\clearpage
\maketitlesupplementary
\appendix
\renewcommand\thefigure{A\arabic{figure}}
\setcounter{figure}{0}
\renewcommand\thetable{A\arabic{table}}
\setcounter{table}{0}
\renewcommand\theequation{A\arabic{equation}}
\setcounter{equation}{0}
\pagenumbering{arabic}
\renewcommand*{\thepage}{A\arabic{page}}
\setcounter{footnote}{0}

\section{Datasets and Benchmarks}
\label{sec:rationale}

\noindent \textbf{Robocasa.} The dataset consists of robot manipulation data extracted from the MuJoCo simulation environment using a Franka Emika Panda robot, with a focus on kitchen scenarios. For our experiments, we used the Human-$50$ (H-$50$) and Generated-$3000$ (G-$3000$) datasets provided by RoboCasa, which are automatically generated using MimicGen \citep{mandlekar2023mimicgen} based on human demonstrations. The benchmark includes $24$ atomic tasks, as detailed in \Cref{table:robocasa-results}.

\noindent \textbf{Metaworld.} MetaWorld is a commonly used benchmark for meta-reinforcement learning and multi-task learning. It consists of 50 distinct robotic manipulation tasks involving a Sawyer robot arm in simulation. The observation is an RGB image of size $64\times64$, and the action is a $4$-dimensional continuous vector.


\begin{table}[h!]
\centering
\caption{Hyper-parameters of the model-based RL experiments.}
\resizebox{\linewidth}{!}{\begin{tabular}{lll}
\toprule
Model-based RL & Hyper-parameter & Value \\
\midrule
\multirow{4}{*}{Rollout Phase} & Init rollout batch size & 640 \\
 & Interval & 200 steps \\
 & Batch size & 32 \\
 & Horizon & 10 \\
\midrule
\multirow{11}{*}{Training phase} & Init training steps & 1000 \\
 & world model training interval & 10 steps \\
 & Batch size & 16 \\
 & Sequence length & 12 \\
 & Context frames & 2 \\
 & Prediction horizon per inference & 1 \\
 & Learning rate & $1 \times 10^{-4}$ \\
 & Optimizer & AdamW \\
\bottomrule
\end{tabular}}
\label{tab:hyper-parameters}
\end{table}

\begin{table}[h]
\centering
\caption{Hyper-parameters of the Imitation Learning experiments.}
\resizebox{0.95\linewidth}{!}{\begin{tabular}{ll}
\toprule
Hyper-parameter & Value \\
\midrule
Policy Embedding dimension & 512 \\
Number of transformer layers & 6 \\
Number of attention heads & 8 \\
Context length & 10 \\
Activation & GELU \\
\midrule
Algorithm & Behavioral Cloning \\
Batch size & 16 \\
Learning rate & 1e-4 \\
Optimizer & AdamW \\
L2 regularization & 0.01 \\
\midrule
Number of atomic tasks & 24 \\
Training data & 50 demos per task \\
Frame stack & 10 \\
\bottomrule
\end{tabular}}
\label{tab:hyper-parameters-il}
\end{table}

\section{Implementation Details}
\label{app:method}
\subsection{EDM Preconditioning}\label{app:method:edm_precond}

As mentioned in Section \ref{subsec:dynamics modeling}, we list the preconditioners here that are designed to improve network training \citep{karras2022elucidating}:

\begin{equation}
    c_{in}^\tau = \frac{1}{\sqrt{\sigma(\tau)^2 + \sigma_{data}^2}}
\end{equation}
\begin{equation}
    c_{out}^\tau = \frac{\sigma(\tau)\sigma_{data}}{\sqrt{\sigma(\tau)^2 + \sigma_{data}^2}}
\end{equation}
\begin{equation}
    c_{noise}^\tau = \frac{1}{4}\log(\sigma(\tau))
\end{equation}
\begin{equation}
    c_{skip}^\tau = \frac{\sigma_{data}^2}{\sigma_{data}^2 + \sigma^2(\tau)},
\end{equation}
where $\sigma_{data}=0.5$.
The noise parameter $\sigma(\tau)$ is sampled to maximize the effectiveness of training as follows:
\begin{equation}
\log(\sigma(\tau))\sim \mathcal{N}(P_{mean}, P_{std}^2),
\end{equation}
where $P_{mean}=-0.4, P_{std}=1.2$.

\subsection{Architectural Design}\label{app:method:model_architecture}

The variational autoencoder employs a transformer-based architecture with point embedding for encoding point cloud inputs. It uses farthest point sampling to downsample the original point cloud ($N=2048$) to a manageable number of latent points ($M=512$), followed by a series of self-attention and cross-attention blocks. For the probabilistic variant, the encoder outputs mean and logvar parameters to sample latent vectors through the reparameterization trick, with an optional KL divergence regularization term. The diffusion model $\mathcal{D}_\theta$ is structured as a Vision Transformer (DiT), processing pointmap patches through multiple transformer blocks with adaptive layer normalization (adaLN) for conditioning on timesteps and actions. The input consists of stacked current, noisy next observations, time embedding, and the current action embedding. The model predicts the denoised next state according to EDM formulation. The reward model $R_\psi$ combines convolutional encoding with sequential modeling, consisting of ResBlocks with optional attention layers followed by an LSTM. The encoder processes pairs of observations (current and next states) while conditioning on embedded actions, and the LSTM captures temporal dependencies before final reward prediction through an MLP head. Before inference, the LSTM hidden states are initialized through a burn-in procedure with conditioning frames.

\subsection{Hyper-parameters}\label{app:method:hyper-parameters}

The hyper-parameters of the Robocasa and Metaworld experiments are listed in \Cref{tab:hyper-parameters-il} and \Cref{tab:hyper-parameters}, respectively.

\end{document}